\documentclass[runningheads]{llncs}

\usepackage{eccv}

\usepackage{eccvabbrv}

\usepackage{graphicx}
\usepackage{booktabs}
\usepackage{multirow}
\usepackage{placeins}
\usepackage{microtype}

\usepackage[accsupp]{axessibility}  

\usepackage[table]{xcolor}
\usepackage[inkscapelatex=false]{svg}
\usepackage{amssymb}
\usepackage{pifont}

\usepackage[breaklinks,colorlinks,citecolor=eccvblue]{hyperref}

\newcommand{\ours}{SegFS\xspace}
\newcommand{\tit}[1]{\smallbreak\noindent\textbf{#1.}}
\newcommand{\tinytit}[1]{\noindent\textbf{#1.}}

\newcommand{\maxim}[1]{\textbf{\color{blue}[Maxim: #1]}}

\newcommand{\marcy}[1]{\textbf{\color{magenta}[Marcella: #1]}}

\newcommand{\luca}[1]{\textbf{\color{green}[Luca: #1]}}
\newcommand{\martin}[1]{\textbf{\color{red}[Martin: #1]}}
\newcommand{\nikita}[1]{\textbf{\color{orange}[Nikita: #1]}}

\renewcommand{\maxim}[1]{}
\renewcommand{\marcy}[1]{}
\renewcommand{\luca}[1]{}
\renewcommand{\martin}[1]{}
\renewcommand{\nikita}[1]{}

\definecolor{blond}{rgb}{0.98, 0.94, 0.75}
\definecolor{OurColor}{RGB}{232, 240, 254}

\newcommand{\cmark}{\ding{51}}%
\newcommand{\xmark}{\ding{55}}%

\makeatletter

\newcommand{\makesupptitle}{%
  \begin{center}%
    \def\lastand{\ifnum\value{@inst}=2\relax
                   \unskip{} \andname\
                 \else
                   \unskip \lastandname\
                 \fi}%
    \def\and{\stepcounter{@auth}\relax
             \ifnum\value{@auth}=\value{@inst}%
               \lastand
             \else
               \unskip,
             \fi}%
    \let\newline\\
    {\Large \bfseries\boldmath
      \pretolerance=10000
      \@title \par}%
    \vskip .3cm
    {\large \bfseries Supplementary Material\par}%
  \end{center}%
}

\makeatother

\usepackage{orcidlink}

\begin{document}

\sloppy

\title{Segmenting, Fast and Slow:\texorpdfstring{\\}{ }%
Real-Time Open-Vocabulary Video Instance Segmentation with Dual-Path Processing}

\titlerunning{Segmenting, Fast and Slow}


\author{Luca~Barsellotti\inst{3}\thanks{Work done during an internship at Google.}\qquad
Martin~Sundermeyer\inst{1}\qquad
Mattia~Segu\inst{1}\qquad %
Nikita~Araslanov\inst{2}\qquad %
Muhammad~Ferjad~Naeem\inst{1}\qquad %
Marcella~Cornia\inst{3}\qquad %
Yongqin~Xian\inst{1}\qquad %
Maxim~Berman\inst{1} %
}

\authorrunning{L.~Barsellotti et al.}


\institute{Google
\and TU Munich; Munich Center for Machine Learning
\and University of Modena and Reggio Emilia}

\maketitle

\vspace{-1em}
\begin{abstract}
Object-centric models inspired by DETR have become the dominant paradigm for open-vocabulary video instance segmentation (OV-VIS). While recent efforts have reduced the computational cost of pixel decoding, textual modality fusion, and object decoding to make these architectures more suitable for mobile devices, real-time on-device inference at high frame rates remains an open challenge. In this paper, we introduce \ours, a dual-stream fast-slow framework that significantly improves efficiency without sacrificing accuracy. On sparse keyframes, an open-vocabulary object-based model predicts instance-level representations. These representations are then projected back into the backbone feature space to condition a lightweight fast network, which efficiently relocalizes and segments the instances in subsequent frames. By shifting instance propagation from object decoding to feature-space conditioning, our approach decouples multimodal semantic understanding from dense mask prediction and enables efficient temporal propagation. The proposed fast branch achieves up to 14× lower latency than the mobile-oriented MOBIUS model, while maintaining competitive segmentation performance on standard OV-VIS benchmarks.

\end{abstract}

\section{Introduction}
\label{sec:intro}

While modern deep learning models exhibit remarkable accuracy and scalability, their substantial computational overhead heavily restricts real-time deployment on edge devices. This bottleneck becomes particularly severe for dense spatio-temporal tasks.
In this work, we focus on Open-Vocabulary Video Instance Segmentation (OV-VIS), which aims at \emph{identifying} and \emph{tracking} object masks of \emph{open-set} categories specified by text prompts.
Established architectures, such as GLEE~\cite{wu2024general}, comprise three main vision components: \emph{(i)} a visual backbone that produces multi-scale features; \emph{(ii)} a feature enhancer that fuses the multi-scale features in a single high-resolution feature map and injects the text embeddings to guide the segmentation process toward identifying corresponding instances; and \emph{(iii)} an object decoder that extracts representative feature vectors describing the identified instances. Among these, the feature enhancer represents the most severe computational bottleneck. 

\begin{figure}[t]
    \centering
     \includegraphics[width=\linewidth]{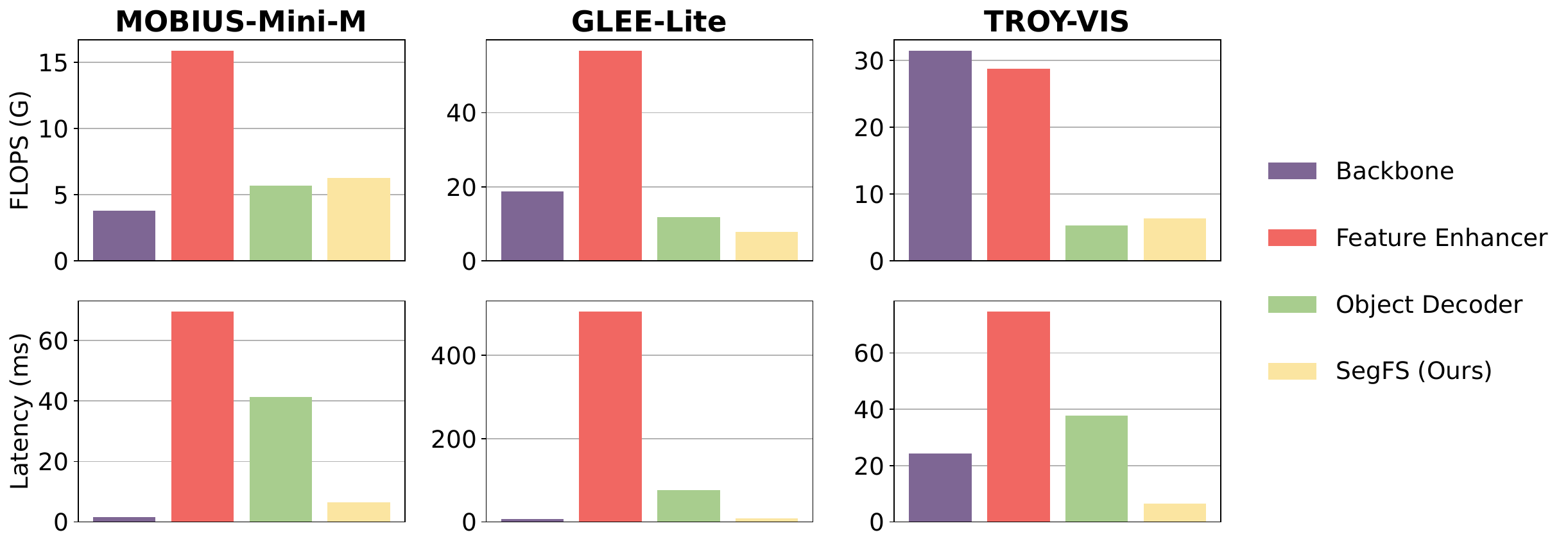}
    \vspace{-0.5cm}
    \caption{Efficiency analysis of OV-VIS architectures. We compare the FLOPs (top) and on-device latency on a Samsung Galaxy S25 Ultra (bottom) for the core components of three baseline models, MOBIUS-Mini-M~\cite{segu2025mobius}, GLEE-Lite~\cite{wu2024general}, and TROY-VIS~\cite{wang2023towards}, when assuming an input image with resolution 480$\times$480 and 40 textual categories, and the Fast Feature Aggregation from our proposed SegFS.}
    \label{fig:flops_and_latency_histogram}
    \vspace{-0.8em}
\end{figure}

Recent works, such as MOBIUS~\cite{segu2025mobius} and TROY-VIS~\cite{wang2023towards}, have introduced lighter and more mobile-friendly feature enhancer modifications.
However, such components still overwhelmingly dominate the inference cost:
\cref{fig:flops_and_latency_histogram} visualizes the FLOPS and the on-device latency on a Samsung Galaxy S25 Ultra of the components of MOBIUS-Mini-M, GLEE-Lite, and TROY-VIS.
This dominance is even more pronounced in on-device latency than in theoretical FLOPs, and scales prohibitively with higher image resolutions and vocabulary size.

To overcome this bottleneck, we draw inspiration from keyframe-based methods that \emph{amortize} computational costs by executing specific network components only periodically.
For example, MobileInst~\cite{Zhang:2024:MobileInst} reuses object embeddings, derived from a keyframe, on intermediate frames, avoiding the overhead of running the object decoder on every frame.
We extend this idea by amortizing the runtime of both the object decoder and feature enhancer.
This approach leads to a highly efficient dual-path architecture: the slow path operates only on the sparse keyframes, whereas the fast path performs lightweight inference on every frame.

Accordingly%
, we introduce \ours
,
a novel dual-stream framework designed for real-time OV-VIS.
While relying on the established models in the slow path, the fast path of SegFS bypasses the computational overhead of the feature enhancer and object decoder.
Our core intuition is that the spatial semantics required for fine-grained localization are already encoded in the feature maps of the visual backbone. 
Therefore, the fast path reuses these feature maps in a lightweight fashion with a novel architecture design.
Meanwhile, on periodic keyframes, the slow path identifies object embeddings and projects them back into the feature space of the backbone, operating exclusively on sparsely sampled keyframes.
\ours effectively injects the projected embeddings into the fast path, progressively fusing the backbone features to produce accurate and temporally consistent segmentation masks.
Extensive experiments demonstrate that our approach achieves up to a 14$\times$ inference speedup over the mobile-friendly object-centric model MOBIUS-Mini-M, delivering $\sim$$3\times$ amortized FPS and crossing the 30~FPS real-time threshold, while preserving accuracy within $\leq$1~AP across standard video segmentation benchmarks relative to the reference upper bound.

Our contributions are as follows:
\begin{itemize}
    \item We extend the keyframe-based paradigm in OV-VIS \cite{Zhang:2024:MobileInst,wang2023towards} to a dual-path approach, where selected network components operate either in the ``fast'' or the ``slow'' path. Specifically, we show that for OV-VIS, backbone features already provide most of the spatial detail needed for localization; the feature enhancer can be moved to sparse keyframes without degrading mask quality.
    \item We implement the dual-path approach in a novel framework, \ours, which effectively leverages the features of the visual backbone in a lightweight fusion module. By injecting instance embeddings into the backbone's feature pyramid, \ours produces highly accurate masks with minimal latency.
    \item Through extensive evaluation on OV-VIS benchmarks, we demonstrate that \ours preserves the zero-shot capabilities and mask quality of heavy object-centric models while achieving
    real-time inference speed.
\end{itemize}

\section{Related Works}
\label{sec:related_works}

\tinytit{Object-Centric Methods} Recent advances in visual recognition increasingly adopt an object-centric formulation, where a fixed set of learnable queries interacts with image features to produce object-level representations. Transformer-based detectors, such as DETR~\cite{carion2020end}, established this paradigm, improved later with more efficient attention and multi-scale feature aggregation~\cite{zhu2021deformable}.
MaskFormer, Mask2Former, and MaskDINO~\cite{cheng2021per,cheng2022masked,li2023mask} extend this formulation to segmentation, where object queries produce instance embeddings interacting with dense feature maps to generate segmentation masks.

More recently, object-centric frameworks have been extended to open-vocabulary recognition by aligning object representations with language embeddings~\cite{zang2022open,liu2024grounding}. Many of these approaches inject textual information directly into the visual feature hierarchy, conditioning dense visual representations with language before object extraction. This early vision-language fusion has been adopted by several open-vocabulary detection and segmentation frameworks~\cite{xu2022simple,ding2022decoupling,liang2023open,zou2023generalized,zou2023segment,zhang2023simple,wu2024general,liang2025referdino}. Our work builds upon these models while addressing their substantial computational overhead.

\tit{Efficient Object-Centric Methods}
The transition from anchor-based CNNs to object-centric, transformer-based architectures like DETR eliminated time-consuming heuristics such as NMS, but introduced new performance bottlenecks such as the quadratic complexity of global attention. 
DeformableDETR~\cite{zhu2021deformable}, SparseDETR~\cite{roh2021sparse} and FocusDETR~\cite{zheng2023less} use sparse attention on a subset of tokens to target this bottleneck. EfficientDETR~\cite{yao2021efficientdetr} leverages dense prior initializations to reduce the required decoder depth. The RT-DETR family~\cite{zhao2024detrs,lv2024rt} applies intra-scale interactions strictly on low-resolution features, decoupling them from cross-scale feature fusion. Grounding DINO 1.5 Edge~\cite{liu2024grounding} extends the RT-DETR decoder to efficient multi-modal fusion for open-set detection. 
GLEE~\cite{wu2024general} unifies instance segmentation tasks across datasets, and MOBIUS~\cite{segu2025mobius} eliminates the redundant multi-scale processing in the transformer decoder of GLEE by condensing multi-scale features into a single representation. However, as can be seen in \cref{fig:flops_and_latency_histogram}, the main bottleneck remains the expensive feature enhancer.

\tit{Temporal Propagation} In videos, temporal feature propagation can further optimize performance and efficiency. 
For encoding frame-by-frame motion, optical flow builds a cornerstone of video understanding.
Classic variational methods, such as Lucas-Kanade and Horn-Schunck \cite{Horn:1981:HS,Lucas:1981:AII}, are computationally heavy and rely on hand-designed regularization.
Modern deep estimators learn strong motion priors, achieving substantially higher throughput \cite{hui20liteflownet2} and accuracy \cite{Ilg:2017:FNE,teed2020raft}.
These properties made optical flow attractive for video segmentation \cite{Sevilla-Lara:2016:OFS,Nilsson:2018:SVS,Xiao:2018:MND}.
For example, MPVSS~\cite{weng2023mask}, compared in our experiments, warps keyframe mask logits to intermediate frames by learning object-conditioned motion fields.
Object-agnostic methods, such as SAM2~\cite{ravisam} and their faster counterparts~\cite{zhang2023mobilesamv2,zhao2023fast}, perform mask propagation of an initial instance segmentation. However, they do not re-use the feature representation of the instance segmenter and require memory banks for long-term tracking.
By contrast, \ours avoids explicit flow estimation and mask propagation, enabling training from still images and reducing on-device computational overhead.

The DETR paradigm naturally extends to videos. MinVIS~\cite{Huang:2022:MinVIS} shows that object embeddings learned for image segmentation remain discriminative across frames, enabling video instance segmentation via simple embedding matching without dedicated spatio-temporal architectures.
MobileInst~\cite{Zhang:2024:MobileInst} proposes a mobile architecture that reuses keyframe object embeddings for segmenting intermediate frames, saving the cost of the object decoder. TROY-VIS~\cite{yan2025towards} extends MobileInst to real-time open-vocabulary VIS on modern GPUs.
Advancing this line of work, \ours moves toward real-time OV-VIS on compute-constrained \emph{mobile} devices.
In particular, \ours circumvents the costly feature enhancer by projecting the slow, semantically rich object tokens into the fast backbone feature space, thereby propagating its high-quality masks while improving efficiency.

\section{Preliminaries}
\label{sec:prelims}

This section introduces the core concepts underlying our dual-path approach. We establish the formal definition of the OV-VIS task, followed by a review of standard and efficient object-centric frameworks that act as the semantic backbone of our slow path.

\tit{Task Definition} We tackle the task of Open-Vocabulary Video Instance Segmentation (OV-VIS). Given an arbitrary set of $M$ free-form textual categories $\mathcal{C} = \{c_1, c_2, \dots, c_M\}$ and an input video sequence $V \in \mathbb{R}^{T \times H \times W \times 3}$ consisting of $T$ frames with spatial resolution $H \times W$, the objective is to simultaneously localize, track, and classify all relevant object instances across the video. Unlike standard closed-set VIS, the categories in $\mathcal{C}$ may contain novel concepts unseen during training. Formally, an OV-VIS model maps the video $V$ and the category set $\mathcal{C}$ to a set of $N$ tracked object instances. For each detected instance $n \in \{1, \dots, N\}$, the model predicts: \textit{(i)} a category label $\hat{y}_n \in \mathcal{C}$, and \textit{(ii)} a temporal sequence of binary segmentation masks $\mathcal{M}_n = \{m_{n,t}\}_{t=1}^T$, where $m_{n,t} \in \{0, 1\}^{H \times W}$ represents the spatial localization of the $n$-th instance at frame $t$. If the object is occluded or absent in a specific frame $t$, the corresponding mask $m_{n,t}$ is empty.
In summary, this problem formulation aims at robust per-frame spatial representations $m_{n,t}$ and consistent temporal associations to form $\mathcal{M}_n$, driven by the semantic guidance of $\mathcal{C}$.

\tit{Open-Vocabulary Object-Centric Frameworks}
Modern object-centric frameworks such as GLEE~\cite{wu2024general}, consist of a visual backbone, a feature enhancer and an object decoder.
Given an input video frame $I \in \mathbb{R}^{H \times W \times 3}$, a standard visual backbone (\eg~ResNet-50) first extracts a multi-scale feature pyramid $\mathcal{F} = \{P_2, P_3, P_4, P_5\}$ with strides of 4, 8, 16, and 32, respectively (\ie $P_2$ has the highest, $P_5$ has the lowest resolution).
The feature enhancer consists of a pixel decoder and an early fusion module~\cite{UNINEXT}, which bridges the visual and textual modalities with bi-directional cross-attention between the visual features and the text embeddings derived from $\mathcal{C}$.
The pixel decoder (implemented with a deformable Transformer) processes these prompt-aware features to exchange information across scales.
Following MaskDINO~\cite{li2023mask}, a selection of $Q$ features from the pixel decoder initialize the queries for the object decoder.
Within the object decoder, these $Q$ queries attend to the multi-scale features from the pixel decoder, resulting in $Q$ updated \emph{object} queries.
For the classification output, these queries are used to compute alignment scores with the text embeddings.
For the segmentation output, these queries play the role of the kernel in a 1$\times$1 convolution with the high-resolution pixel embedding map $M_{p}$ (at $1/4$ resolution) resulting from the feature enhancer.
The result of the final convolution are mask logits, which are thresholded to yield the binary object-specific segmentation masks.
Finally, to extend this per-frame processing to the video domain, these architectures employ the MinVIS~\cite{Huang:2022:MinVIS} tracking-by-matching paradigm, efficiently associating object queries and their corresponding masks across consecutive frames by computing the optimal bipartite matching over the cosine similarity of their instance embeddings.

\tit{Efficient Methods} 
The standard object-centric framework, described above, builds the foundation for state-of-the-art accuracy.
However, the dense interactions within the early fusion and multi-scale pixel decoder result in a severe computational bottleneck. 
Recent works, such as MOBIUS~\cite{segu2025mobius} and TROY-VIS \cite{wang2023towards}, introduce structural modifications to these modules.
Specifically, MOBIUS \cite{segu2025mobius} selects a single intermediate scale in the pixel decoder to perform the early fusion with the text modality and multi-scale attention, and to guide the object decoding, significantly improving the efficiency of the model.
TROY-VIS \cite{wang2023towards} accelerates the early fusion module by decomposing the feature enhancer into two distinct lightweight operations, a vision-language cross-attention to the lowest-resolution visual feature map and a scale attention module that propagates these text-conditioned semantics bottom-up across the higher-resolution scales.

\nikita{What's missing in this section is 'why' -- why do we explain this all in such detail? Are we using / extending / building on these techniques? This needs to be explained.} \maxim{Agreed, the secion is also confusing because it's not claer what exactly the link and differentiating factor w.r.t. prior art is; if we use something standard, it should be very short to mention it with a reference; if we modify something, it should be clear in what way and why. Let's improve the flow between this section and the next one -- merging it into one if needed}

\section{\ours: Segmenting, Fast and Slow}
\label{sec:method}

\begin{figure}[!t]
    \centering
    \includegraphics[width=\linewidth]{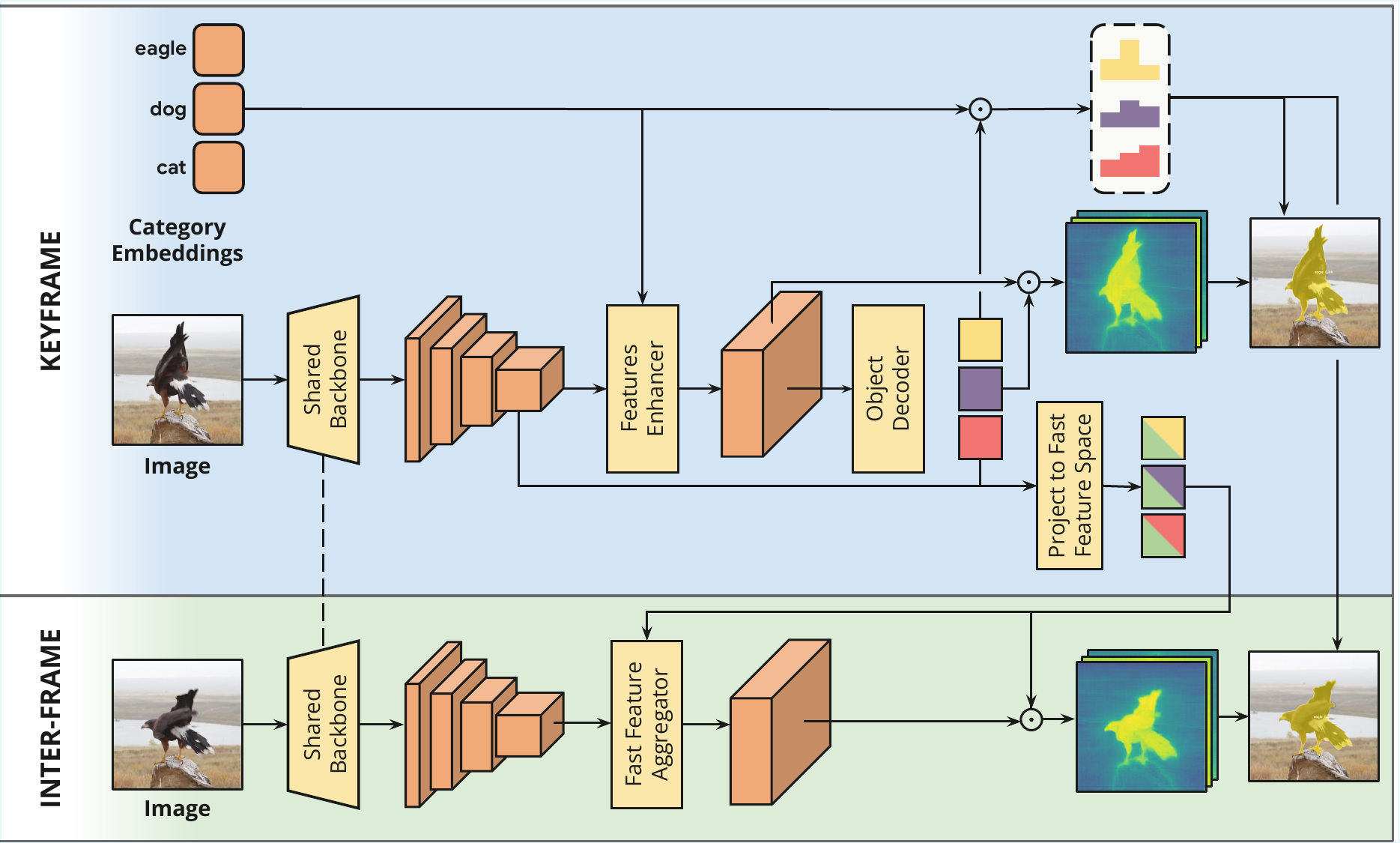}
    \vspace{-0.4cm}
    \caption{Overview of our SegFS architecture. The slow path (top) extracts object embeddings from sparse keyframes to condition the lightweight fast path (bottom) on intermediate frames.}
    \label{fig:dual_stream}
    \vspace{-0.6cm}
\end{figure}

\tinytit{Overview} 
To overcome the persistent latency bottlenecks of existing frameworks, we propose \ours, a novel dual-stream architecture for real-time OV-VIS.
Our approach is based on the strategic temporal alternation between two distinct networks, as shown in~\cref{fig:dual_stream}.
The ``slow'' network, represented by  GLEE \cite{wu2024general}, MOBIUS \cite{segu2025mobius}, or TROY-VIS \cite{wang2023towards} in our experiments, is a frozen object-centric OV-VIS model, described in \cref{sec:prelims}.
Operating exclusively on sparsely sampled keyframes, the slow network extracts text-aligned object embeddings, propagated and reused across intermediate frames.
For the intermediate $T$ frames (or inter-frames), we develop a lightweight ``fast'' network, designed to rapidly generate a spatial feature map.
The fast network bypasses the computationally expensive decoding stages and relies on the multi-scale features from the backbone, enhanced by fusing the object embeddings from the keyframes.
Finally, we reuse these keyframe-derived object embeddings for temporal association across the entire video sequence using the standard MinVIS tracking-by-matching paradigm.

\tit{Projecting Object Embeddings to Fast Feature Space}
Our fast network reuses the object embeddings decoded from the slow network on the keyframes. The core intuition is that, once the instances are effectively detected by the slow network with the computationally heavy feature enhancer and object decoder, their embeddings can be projected into the backbone feature space while still preserving instance discrimination.
This projection is done by a 3-layer feed-forward network (FFN) followed by a LayerNorm.
Next, the network splits into two branches. 
The first branch generates the tokens that will condition the inference in the fast network. 
Here, we select only the top $K$ object embeddings based on their maximum similarity scores with the text categories.
This prevents the injection of noisy or irrelevant queries into the fast network.
We then append a learnable background token, $t_{bg}$, to explicitly model non-object regions, resulting in $K+1$ tokens.
These tokens are processed through two interleaved blocks of self-attention and cross-attention.
In the cross-attention layers, the $K+1$ tokens act as queries; the feature map, $P_5$, from the backbone run on the keyframe serves as the keys and values.
Recall that $P_5$ has the lowest resolution and the highest degree of semantic content in the feature pyramid $P_i$.
In parallel, the second branch applies another 3-layer FFN to project the embeddings into the kernel space for 1$\times$1 convolution with the mask logits.

\begin{figure}[!t]
    \centering
    \includegraphics[width=\linewidth]{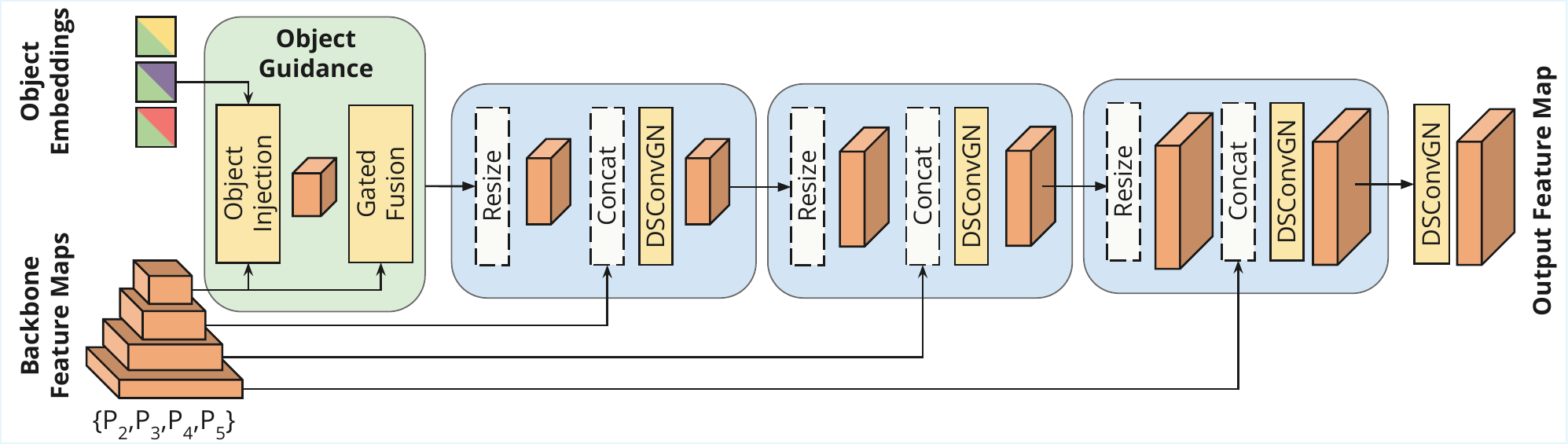}
    \vspace{-0.5cm}
    \caption{Overview of our Fast Feature Aggregator. Keyframe object embeddings are injected into the $P_5$ backbone features via Object Guidance, then progressively upsampled and fused until the final feature map.}
    \label{fig:architecture}
    \vspace{-0.2cm}
\end{figure}

\begin{figure}[!t]
    \centering
    \includegraphics[width=\linewidth]{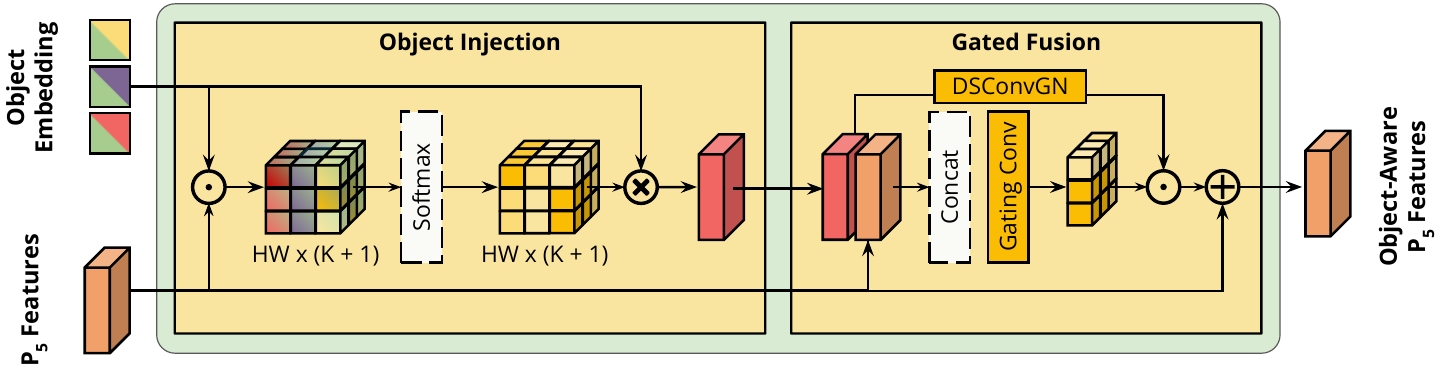}
        \vspace{-0.1cm}
    \caption{Overview of our Object Guidance module. Object embeddings projected from the slow network on a keyframe are spatially injected into $P_5$ and merged using delta-gated fusion to produce an updated object-aware formulation of $P_5$.}
    \label{fig:object_guidance}
    \vspace{-0.6cm}
\end{figure}

\subsection{Fast Feature Aggregator}
To process the $T$ inter-frames with minimal latency, we introduce an efficient Fast Feature Aggregator.
Its goal is to compute high-resolution feature maps conditioned on the object embeddings from the slow path.
\cref{fig:architecture} provides a detailed overview of its architecture.

\tit{Preprocessing} We first project the multi-scale backbone feature maps $\mathcal{F} = \{P_2, P_3, P_4, P_5\}$ to a unified channel dimension ($D=256$).
To maximize efficiency, we apply standard dense 1$\times$1 convolutions to the lower-resolution maps $\{P_4, P_5\}$.
For the high-resolution maps $\{P_2, P_3\}$, we utilize a lightweight mobile-friendly DSConvGN block, consisting of 3$\times$3 depthwise convolution, 1$\times$1 pointwise convolution, GroupNorm, and SiLU.
This preprocessing unifies the channel space, rendering the computational cost of the fast network largely independent of the backbone capacity.

\tit{Object Guidance}
The core of our conditioning mechanism occurs exclusively at the lowest-resolution feature map, $P_5$ (see \cref{fig:architecture}).
Our goal is to ``inject'' $K+1$ object embeddings, extracted from the keyframe, into the spatial cells of $P_5$.
\cref{fig:object_guidance} offers a detailed visualization of the Object Guidance module, which achieves this goal.
It comprises two blocks: Object Injection and Gated Fusion.
In Object Injection, we compute the multi-head cosine similarity between each spatial cell in $P_5$ and the $K+1$ tokens.
These similarities are scaled by a learnable temperature parameter $\tau$ (initialized per-head) and passed through a softmax function across the instance dimension.
By computing a weighted sum of the instance embeddings based on these softmax probabilities, we obtain the object-aware feature map $I_5$.
Notably, as the network learns to lower the temperature $\tau$, the softmax distribution sharpens.
In the limit, each spatial cell in $I_5$ is effectively replaced by its most semantically similar object embedding.

Since the map $I_5$ represents coarse semantic structures, it must be smoothed and grounded back into the current frame's spatial structure.
The Gated Fusion first applies a DSConvGN block to $I_5$ for local smoothing.
Next, we fuse it with the original projected backbone map $P_5$ via a gated mechanism:

\begin{equation}
    \tilde{P}_5 = P_5 + \sigma\Big(\text{GateConv}(P_5 \parallel I_5)\Big) \odot \text{DSConvGN}(I_5),
\end{equation}
where $\parallel$ denotes concatenation, $\sigma$ is the sigmoid activation, and the gate convolution predicts a spatial blending mask.
This ensures the network retains the precise visual cues of $P_5$ while absorbing the instance semantics of $I_5$.

\tit{Progressive Upsampling} 
To refine the boundaries of the localized objects without incurring heavy computational costs, we avoid running the Object Guidance at higher resolutions. 
Instead, we progressively upsample the enriched $\tilde{P}_5$ map, concatenate it with the subsequent backbone feature map (\eg, $P_4$), and fuse both using the lightweight DSConvGN blocks (see \cref{fig:architecture}).
This upsample-and-fuse approach continues until we reach the $P_2$ resolution.
Following the final DSConvGN refinement, this high-resolution feature map feeds to a 1$\times$1 convolution, where the object embeddings from the keyframe play the role of the convolutional kernels, producing $N$ mask activations.

\subsection{Training Strategy}
We train \ours on \emph{image-based} instance segmentation datasets, allowing us to leverage the vast diversity of spatial annotations without requiring expensive video ground truth.
During a training iteration, an input image is processed by the fast and the slow paths sequentially.
First, the slow network executes its standard forward pass to extract the object queries, category logits, and bounding box predictions.
The object queries are then projected to the fast feature space and fed into the fast network.
The fast network runs on top of the same image backbone features to predict the corresponding instance masks.
For the bipartite Hungarian matching with the ground-truth instance masks and categories, we construct a hybrid matching cost: we utilize the highly accurate category logits and bounding box predictions from the pre-trained slow network, while incorporating the mask predictions from the fast network.
By anchoring the matching process with the slow network, we ensure stable and consistent ground-truth assignment.
Once the optimal assignment is determined, the fast network is supervised by the Mask and the DICE losses.
\section{Experiments}
\vspace{-2mm}
\label{sec:experiments}

\begin{table}[!t]
  \caption{Performance comparison of slow and fast models on the YouTubeVIS19~\cite{yang2019video}, OVIS~\cite{qi2022occluded}, BURST~\cite{athar2023burst}, and LV-VIS~\cite{wang2023towards} datasets. The last column reports the amortized FPS on a span of 6 frames when running the models on the Samsung Galaxy S25 Ultra.}
  \vspace{-0.15cm}
  \label{tab:model_comparison}
  \centering
  \small 
  \setlength{\tabcolsep}{.28em}
  \resizebox{\textwidth}{!}{%
  \begin{tabular}{l ccc ccc cccccc cccc}
    \toprule
    & \multicolumn{3}{c}{\multirow{2}{*}{\textbf{YouTubeVIS19}}} & \multicolumn{3}{c}{\multirow{2}{*}{\textbf{OVIS}}} & \multicolumn{6}{c}{\textbf{BURST}} & \multicolumn{3}{c}{\multirow{2}{*}{\textbf{LV-VIS}}}\\
    \cmidrule(lr){8-13} 
     & & & & & & & \multicolumn{2}{c}{All} & \multicolumn{2}{c}{Common} & \multicolumn{2}{c}{Uncommon} & & & \\
     \cmidrule(lr){2-4} \cmidrule(lr){5-7} \cmidrule(lr){8-9} \cmidrule(lr){10-11} \cmidrule(lr){12-13} \cmidrule(lr){14-16}
    \textbf{Fast Model} & AP & AP$_{50}$ & AP$_{75}$ & AP & AP$_{50}$ & AP$_{75}$ & HOTA & mAP & HOTA & mAP & HOTA & mAP & AP & AP$_{b}$ & AP$_{n}$ & \textbf{FPS} \\
    \midrule
    \rowcolor{gray!15}\multicolumn{17}{l}{\textbf{Slow Model:} MOBIUS~\cite{segu2025mobius} \quad \textbf{Backbone:} MNv4-CM} \\
    - & 48.7 & 70.4 & 51.9 & 23.6 & 45.5 & 22.2 & 19.7 & 9.1 & 40.1 & 16.0 & 14.6 & 7.4 & 16.7 & 21.0 & 13.5 & 8.7\\
    \textit{Copy} & 20.1 & 46.5 & 13.1 & 5.1 & 13.8 & 2.5 & 11.4 & 2.9 & 22.7 & 3.2 & 8.6 & 2.8 & 10.4 & 12.8 & 8.7 & 51.8\\
    \textit{Reuse Objects} & 42.1 & 63.8 & 42.6 & 15.0 & 31.4 & 13.0 & 15.2 & 5.9 & 30.6 & 9.2 & 11.3 & 5.1 & 15.8 & 20.1 & 12.7 & 12.7 \\
    \midrule
    LiteFlowNet2~\cite{hui20liteflownet2} & 28.6 & 55.1 & 23.6 & 8.0 & 22.3 & 4.9 & 12.8 & 3.5 & 25.6 & 5.4 & 9.6 & 3.0 & 12.2 & 14.8 & 10.3 & 21.6\\
    RAFT~\cite{teed2020raft} & 28.9 & 56.3 & 25.0 & 8.0 & 21.8 & 5.0 & 12.5 & 3.4 & 25.2 & 5.1 & 9.3 & 3.0 & 12.1 & 14.7 & 10.3 & 10.9 \\
    MPVSS~\cite{weng2023mask} & 39.8 & 63.6 & 39.5 & 13.9 & \textbf{31.9} & \textbf{11.6} & 13.8 & 5.2 & 28.4 & 7.6 & 10.2 & 4.6 & 14.8 & 18.5 & 12.1 & 16.5 \\
    \rowcolor{OurColor}
    \ours (Ours) & \textbf{41.6} & \textbf{63.8} & \textbf{44.6} & \textbf{14.1} & 30.6 & 11.3 & \textbf{14.7} & \textbf{5.9} & \textbf{30.3} & \textbf{9.4} & \textbf{10.8} & \textbf{5.0} & \textbf{15.2} & \textbf{19.2} & \textbf{12.3} & \textbf{38.2}\\
    \rowcolor{OurColor} \textbf{$\Delta$ SegFS - Reuse Obj.} & {\color{BrickRed}-0.5} & {\color{gray}0.0} & {\color{ForestGreen}+2.0} & {\color{BrickRed}-0.9} & {\color{BrickRed}-0.8} & {\color{BrickRed}-1.7} & {\color{BrickRed}-0.5} & {\color{gray}0.0} & {\color{BrickRed}-0.3} & {\color{ForestGreen}+0.2} & {\color{BrickRed}-0.5} & {\color{BrickRed}-0.1} & {\color{BrickRed}-0.6} & {\color{BrickRed}-0.9} & {\color{BrickRed}-0.4} & {\color{ForestGreen}+25.5}\\
    \midrule
    \rowcolor{gray!15} \multicolumn{17}{l}{\textbf{Slow Model:} MOBIUS~\cite{segu2025mobius} \quad \textbf{Backbone:} MNv4-CL} \\
    - & 51.0 & 73.0 & 56.1 & 27.3 & 49.9 & 26.4 & 23.4 & 11.2 & 41.4 & 18.4 & 19.0 & 9.4 & 20.1 & 24.6 & 16.7 & 8.6 \\
    \textit{Copy} & 21.6 & 48.2 & 15.0 & 7.1 & 18.2 & 4.0 & 13.0 & 3.3 & 22.3 & 3.4 & 10.7 & 3.3 & 12.6 & 14.9 & 10.9 & 51.8 \\
    \textit{Reuse Objects} & 44.9 & 68.1 & 46.1 & 18.3 & 35.3 & 16.9 & 17.4 & 6.9 & 31.5 & 10.2 & 13.9 & 6.1 & 18.8 & 22.8 & 15.8 & 12.5 \\
    \midrule
    LiteFlowNet2~\cite{hui20liteflownet2} & 30.7 & 58.0 & 26.7 & 10.6 & 25.8 & 8.0 & 14.6 & 4.0 & 25.8 & 5.3 & 11.8 & 3.7 & 14.6 & 16.9 & 13.0 & 21.6 \\
    RAFT~\cite{teed2020raft} & 31.0 & 59.5 & 27.0 & 10.6 & 25.9 & 7.7 & 14.3 & 3.8 & 25.3 & 5.1 & 11.6 & 3.5 & 14.6 & 16.8 & 13.0 & 10.9 \\
    MPVSS~\cite{weng2023mask} & 42.5 & 66.5 & 45.2 & 16.8 & 33.5 & 15.0 & 15.3 & 6.3 & 27.9 & 8.4 & 12.2 & 5.7 & 17.6 & 21.0 & 15.1 & 16.5 \\
    \rowcolor{OurColor}
    \ours (Ours) & \textbf{44.9} & \textbf{67.7} & \textbf{49.6} & \textbf{17.3} & \textbf{34.1} & \textbf{16.2} & \textbf{16.6} & \textbf{6.9} & \textbf{30.5} & \textbf{10.9} & \textbf{13.1} & \textbf{5.9} & \textbf{18.1} & \textbf{21.7} & \textbf{15.4} & \textbf{36.2}\\
    \rowcolor{OurColor} \textbf{$\Delta$ SegFS - Reuse Obj.} & {\color{gray}0.0} & {\color{BrickRed}-0.4} & {\color{ForestGreen}+3.7} & {\color{BrickRed}-1.0} & {\color{BrickRed}-1.2} & {\color{BrickRed}-0.7} & {\color{BrickRed}-0.8} & {\color{gray}0.0} & {\color{BrickRed}-1.0} & {\color{ForestGreen}+0.7} & {\color{BrickRed}-0.8} & {\color{BrickRed}-0.2} & {\color{BrickRed}-0.7} & {\color{BrickRed}-1.1} & {\color{BrickRed}-0.4} & {\color{ForestGreen}+24.7}\\
    \midrule
    \rowcolor{gray!15} \multicolumn{17}{l}{\textbf{Slow Model:} MOBIUS~\cite{segu2025mobius} \quad \textbf{Backbone:} ResNet50} \\
     - & 56.4 & 78.4 & 59.9 & 30.6 & 51.7 & 30.6 & 13.5 & 12.4 & 43.9 & 19.6 & 18.4 & 10.6 & 19.1 & 23.1 & 16.2 & 8.0 \\
    \textit{Copy} & 23.6 & 50.3 & 16.2 & 7.0 & 18.3 & 3.7 & 12.8 & 2.3 & 23.4 & 3.5 & 10.2 & 2.3 & 11.6 & 13.7 & 10.0 & 48.2 \\
    \textit{Reuse Objects} & 50.3 & 71.8 & 54.7 & 20.3 & 39.3 & 18.6 & 18.0 & 7.3 & 32.9 & 11.4 & 14.3 & 6.3 & 17.8 & 21.6 & 14.9 & 11.2 \\
    \midrule
    LiteFlowNet2~\cite{hui20liteflownet2} & 32.8 & 59.9 & 29.8 & 10.8 & 26.3 & 8.2 & 14.8 & 3.7 & 26.9 & 5.5 & 11.8 & 3.2 & 13.6 & 15.9 & 11.9 & 20.9\\
    RAFT~\cite{teed2020raft} & 32.9 & 59.5 & 29.7 & 10.8 & 26.0 & 8.6 & 14.4 & 3.6 & 26.4 & 5.3 & 11.5 & 3.2 & 13.6 & 15.9 & 11.9 & 10.8 \\
    MPVSS~\cite{weng2023mask} & 44.6 & 68.3 & 47.4 & \textbf{17.3} & \textbf{36.9} & \textbf{15.6} & 15.3 & 5.1 & 28.3 & 7.6 & 12.1 & 4.5 & \textbf{16.3} & \textbf{19.5} & 14.0 & 16.1 \\
    \rowcolor{OurColor}
    \textbf{\ours} (Ours) & \textbf{45.0} & \textbf{69.9} & \textbf{49.0} & 15.7 & 34.7 & 13.9 & \textbf{16.1} & \textbf{6.4} & \textbf{29.4} & \textbf{10.0} & \textbf{12.8} & \textbf{5.5} & \textbf{16.3} & 19.3 & \textbf{14.2} & \textbf{30.2} \\
    \rowcolor{OurColor} \textbf{$\Delta$ SegFS - Reuse Obj.} & {\color{BrickRed}-5.3} & {\color{BrickRed}-1.9} & {\color{BrickRed}-5.7} & {\color{BrickRed}-4.6} & {\color{BrickRed}-4.6} & {\color{BrickRed}-4.7} & {\color{BrickRed}-1.9} & {\color{BrickRed}-0.9} & {\color{BrickRed}-3.5} & {\color{BrickRed}-1.4} & {\color{BrickRed}-1.5} & {\color{BrickRed}-0.8} & {\color{BrickRed}-1.5} & {\color{BrickRed}-2.3} & {\color{BrickRed}-0.7} & {\color{ForestGreen}+19.0}\\
    \midrule
    \rowcolor{gray!15} \multicolumn{17}{l}{\textbf{Slow Model:} GLEE~\cite{wu2024general} \quad \textbf{Backbone:} ResNet50} \\
    - & 52.9 & 73.8 & 58.9 & 25.1 & 45.6 & 23.9 & 23.3 & 11.3 & 42.9 & 19.4 & 18.4 & 9.3 & 19.2 & 23.3 & 16.1 & 1.7 \\
    \textit{Copy} & 21.3 & 47.7 & 13.4 & 6.2 & 15.5 & 4.2 & 12.7 & 3.1 & 24.0 & 3.8 & 9.9 & 3.0 & 11.7 & 14.3 & 9.8 & 10.2 \\
    \textit{Reuse Objects} & 47.2 & 66.3 & 52.2 & 17.6 & 36.0 & 15.0 & 18.0 & 7.3 & 32.9 & 11.4 & 14.3 & 6.3 & 17.8 & 21.6 & 14.9 & 1.9 \\
    \midrule
    LiteFlowNet2~\cite{hui20liteflownet2} & 32.8 & 59.9 & 29.8 & 10.8 & 26.3 & 8.2 & 14.8 & 3.7 & 26.9 & 5.5 & 11.9 & 3.2 & 13.6 & 15.9 & 11.9 & 8.0 \\
    RAFT~\cite{teed2020raft} & 32.9 & 59.5 & 29.7 & 10.8 & 26 & 8.6 & 14.4 & 3.6 & 26.4 & 5.3 & 11.5 & 3.2 & 13.6 & 15.9 & 11.9 & 7.2 \\
    MPVSS~\cite{weng2023mask} & \textbf{42.5} & \textbf{64.7} & \textbf{46.3} & \textbf{15.2} & \textbf{33.1} & \textbf{12.0} & 15.6 & 5.5 & 29.3 & 8.3 & 12.2 & 4.8 & \textbf{16.3} & \textbf{19.9} & \textbf{13.7} & 5.9 \\
    \rowcolor{OurColor}
    \ours (Ours) & 41.8 & 64.5 & 45.7 & 13.6 & 30.7 & 11.9 & \textbf{16.1} & \textbf{6.3} & \textbf{29.8} & \textbf{9.9} & \textbf{12.7} & \textbf{5.5} & 16.1 & 19.6 & 13.6 & \textbf{9.1}\\
    \rowcolor{OurColor} \textbf{$\Delta$ SegFS - Reuse Obj.} & {\color{BrickRed}-5.4} & {\color{BrickRed}-1.8} & {\color{BrickRed}-4.5} & {\color{BrickRed}-4.0} & {\color{BrickRed}-5.3} & {\color{BrickRed}-3.1} & {\color{BrickRed}-1.9} & {\color{BrickRed}-1.0} & {\color{BrickRed}-3.1} & {\color{BrickRed}-1.5} & {\color{BrickRed}-1.6} & {\color{BrickRed}-0.8} & {\color{BrickRed}-1.7} & {\color{BrickRed}-2.0} & {\color{BrickRed}-1.3} & {\color{ForestGreen}+7.2}\\
    \midrule
    \rowcolor{gray!15} \multicolumn{17}{l}{\textbf{Slow Model:} TROY-VIS~\cite{yan2025towards} \quad \textbf{Backbone:} EfficientViT-L2} \\
    - & 57.4 & 81.7 & 63.3 & 28.0 & 50.3 & 25.8 & 22.0 & 11.2 & 42.8 & 19.2 & 16.8 & 9.2 & 20.8 & 24.4 & 18.1 & 7.4\\
    \textit{Copy} & 24.6 & 54.8 & 15.4 & 6.4 & 16.3 & 4.2 & 12.8 & 2.9 & 24.8 & 3.7 & 9.8 & 2.7 & 12.1 & 14.2 & 10.5 & 44.4\\
    \textit{Reuse Objects} & 52.1 & 74.4 & 59.3 & 16.6 & 32.8 & 15.3 & 17.0 & 6.4 & 33.2 & 10.1 & 13.0 & 5.5 & 18.9 & 22.1 & 16.5 & 9.5 \\
    \midrule
    LiteFlowNet2~\cite{hui20liteflownet2} & 35.0 & 63.6 & 32.6 & 9.1 & 22.5 & 6.4 & 14.7 & 3.6 & 28.0 & 5.6 & 11.4 & 3.1 & 14.4 & 16.2 & 13.0 & 20.2\\
    RAFT~\cite{teed2020raft} & 35.1 & 63.8 & 33.8 & 9.1 & 21.9 & 6.5 & 14.4 & 3.5 & 27.5 & 5.3 & 11.1 & 3.0 & 14.4 & 16.2 & 13.0 & 15.7 \\
    MPVSS~\cite{weng2023mask} & 47.2 & 72.6 & 51.4 & 14.7 & \textbf{31.0} & 13.6 & 15.6 & 5.7 & 30.1 & 8.1 & 12.0 & 5.1 & 17.7 & 20.7 & 15.6 & 10.6 \\
    \rowcolor{OurColor}
    \ours (Ours) & \textbf{50.7} & \textbf{74.7} & \textbf{56.0} & \textbf{14.8} & 29.9 & \textbf{14.4} & \textbf{16.8} & \textbf{6.5} & \textbf{33.0} & \textbf{10.1} & \textbf{12.8} & \textbf{5.6} & \textbf{18.3} & \textbf{21.2} & \textbf{16.3} & \textbf{23.3} \\
    \rowcolor{OurColor} \textbf{$\Delta$ SegFS - Reuse Obj.} & {\color{BrickRed}-1.4} & {\color{ForestGreen}+0.3} & {\color{BrickRed}-3.3} & {\color{BrickRed}-1.8} & {\color{BrickRed}-2.9} & {\color{BrickRed}-0.9} & {\color{BrickRed}-0.2} & {\color{ForestGreen}+0.1} & {\color{BrickRed}-0.2} & {\color{gray}0.0} & {\color{BrickRed}-0.2} & {\color{ForestGreen}+0.1} & {\color{BrickRed}-0.6} & {\color{BrickRed}-0.9} & {\color{BrickRed}-0.2} & {\color{ForestGreen}+13.8}\\
    \bottomrule
  \end{tabular}%
  }
\vspace{-0.48cm}
\end{table}

\begin{table}[!t]
  \caption{Efficiency and latency comparison. Latency is measured at $480 \times 480$ resolution with 40 categories. \ours metrics include both the frozen backbone, shared with the slow models, and the Fast Feature Aggregator.}
  \vspace{-0.15cm}
  \label{tab:efficiency}
  \centering
  \small 
  \setlength{\tabcolsep}{.2em}
  \resizebox{\textwidth}{!}{%
  \begin{tabular}{lc cc cccc cc}
    \toprule
    & & \multicolumn{2}{c}{\textbf{Efficiency}} & \multicolumn{4}{c}{\textbf{Edge Latency (ms)}} & \multicolumn{2}{c}{\textbf{GPU Latency (ms)}} \\
    \cmidrule(lr){3-4} \cmidrule(lr){5-8} \cmidrule(lr){9-10}
    \textbf{Model} & \textbf{Backbone} & FLOPs (G) & Param (M) & S25 & Snap 8 Gen 5 & Snap X2 & XR2 & NVIDIA A100 & NVIDIA T4 \\
    \midrule
    \rowcolor{gray!15}
    MOBIUS~\cite{segu2025mobius} & MNv4-CM & 25.4 & 26.2 & 115.7 & 99.5 & 99.3 & 242.2 & 64.8 & 68.4 \\
    \rowcolor{gray!15}
    MOBIUS~\cite{segu2025mobius} & MNv4-CL & 32.6 & 49.1 & 115.9 & 99.8 & 99.5 & 248.8 & 67.5 & 78.1 \\
    \rowcolor{gray!15}
    MOBIUS~\cite{segu2025mobius} & ResNet50 & 54.2 & 60.6 & 124.5 & 102.5 & 97.9 & - & 79.8 & 92.5 \\
    \rowcolor{gray!15}
    GLEE~\cite{wu2024general} & ResNet50 & 87.5 & 55.5 & 587.9 & 520.4 & 567.9 & - & 76.5 & 132.8 \\
    \rowcolor{gray!15}
    TROY-VIS~\cite{yan2025towards} & EffViT-L2 & 65.7 & 63.7 & 135.2 & 112.3 & 114.6 & 312.0 & 49.4 & 79.9 \\
    \midrule
    LiteFlowNet2~\cite{hui20liteflownet2} & - & 38.8 & 6.4 & 32.4 & 28.3 & 29.2 & 71.7 & 24.7 & - \\
    RAFT~\cite{teed2020raft} & - & 61.8 & \textbf{5.3} & 86.5 & 71.1 & 66.0 & 213.9 & 14.9 & 39.4 \\
    MPVSS~\cite{weng2023mask} & - & 39.2 & 40.1 & 49.6 & 45.0 & 43.2 & 159.3 & 12.4 & 42.5 \\
    \midrule
    \rowcolor{OurColor}
    \ours (Ours) & MNv4-CM & \textbf{10.1} & 10.4 & \textbf{8.3} & \textbf{6.8} & \textbf{6.8} & \textbf{21.5} & 13.2 & \textbf{16.1} \\
    \rowcolor{OurColor}
    \ours (Ours) & MNv4-CL & 17.4 & 33.3 & 10.0 & 7.9 & 8.4 & 29.8 & 15.8 & 20.9 \\
    \rowcolor{OurColor}
    \ours (Ours) & ResNet50 & 26.5 & 27.3 & 14.8 & 12.0 & 11.9 & 44.1 & \textbf{9.7} & 22.9 \\
    \rowcolor{OurColor}%
    \ours (Ours) & EffViT-L2 & 37.9 & 52.2 & 31.0 & 28.1 & 28.6 & 91.2 & 17.4 & 32.6 \\

    \bottomrule
  \end{tabular}%
  }
\vspace{-0.6cm}
\end{table}

\subsection{Experimental Setup}
\label{sec:setup}
\tinytit{Implementation Details} We follow the unified training stage proposed in~\cite{segu2025mobius} and train our models \textit{(i)} on the image instance segmentation datasets COCO~\cite{lin2014microsoft}, LVIS~\cite{gupta2019lvis}, and BDD~\cite{yu2020bdd100k}; \textit{(ii)} on the video instance segmentation datasets YouTubeVIS19, YouTubeVIS21~\cite{yang2019video}, and OVIS~\cite{qi2022occluded}, treating them as image datasets; \textit{(iii)} on the referring segmentation datasets RefCOCO, RefCOCO+, RefCOCOg~\cite{nagaraja2016modeling}, and RVOS~\cite{seo2020urvos}; and \textit{(iv)} on the open-world segmentation datasets UVO~\cite{wang2021unidentified} and SA-1B~\cite{kirillov2023segment} using the category name \textit{object}. 
As \ours is designed for the instance segmentation task, we excluded object detection datasets, such as Objects365~\cite{shao2019objects365} and OpenImages~\cite{krylov2021open}.
We train our models with multi-scale training on 4 A100 GPUs with a batch size of 128 for 500,000 iterations and learning rate 1e-4.
We keep the slow models frozen and output 300 object queries. We select the top-50 object queries according to the similarity with the text categories to perform the Object Guidance.

\tit{Evaluation Protocol} Following \cite{wu2024general}, we evaluate on two video instance segmentation datasets seen during training: YouTubeVIS19~\cite{yang2019video} (2,883 videos, 40 categories) and OVIS~\cite{qi2022occluded} (901 videos, 25 categories, severe occlusions).
We also evaluate zero-shot performance on two open-world large-vocabulary datasets: BURST~\cite{athar2023burst} (2,907 videos, 425 base/57 novel categories) and LV-VIS~\cite{wang2023towards} (4,828 videos, 1,196 categories). For all experiments, we resize input videos to a short side of 480 pixels, use $T=5$ frames for the mask propagation, and report the standard evaluation metrics associated with each respective dataset.

\subsection{Comparison with the State of the Art}
\label{sec:comparison}

As shown in \cref{tab:model_comparison}, we evaluate the performance of our proposed method against several baselines by utilizing five different object-centric OV-VIS models as the frozen slow network. Specifically, we employ three variants of MOBIUS~\cite{segu2025mobius} (equipped with MobileNetV4-CM~\cite{qin2024mobilenetv4}, MobileNetV4-CL~\cite{qin2024mobilenetv4}, and ResNet50~\cite{he2016deep} backbones), GLEE-Lite~\cite{wu2024general} (ResNet50~\cite{he2016deep}), and TROY-VIS~\cite{yan2025towards} (EfficientViT-L2~\cite{cai2023efficientvit}). Each distinct block within the table represents the results obtained when pairing one of these slow networks with a specific fast-path solution. The first row of each block reports the upper-bound performance achieved by running the full slow network on every frame of the sequence. The last column reports the amortized FPS: they are computed considering the average FPS on a span of 6 frames, containing the keyframe, on which runs the full slow model, and $T=5$ inter-frames, on which runs the fast path, on the Samsung Galaxy S25 Ultra device in the Qualcomm AI Hub (see \textit{Efficiency Analysis} for more details).

\tit{Copy and Reuse Objects} We introduce two straightforward baselines to establish performance bounds: ``Copy'' and ``Reuse Objects''. The ``Copy'' baseline, originally introduced in MPVSS~\cite{weng2023mask}, serves as a basic lower bound. It simply propagates the segmentation masks predicted by the slow network on the keyframe across the subsequent $T$ frames without any modification. While this operation incurs zero computational cost, the static masks completely fail to follow moving objects in the video. The ``Reuse Objects'' baseline is inspired by efficient architectures like MobileInst~\cite{Zhang:2024:MobileInst} and TROY-VIS~\cite{yan2025towards}. In this approach, we execute the slow network on the propagation frames while skipping the object decoder entirely.
We compute the dot product between the multi-scale feature maps produced by the Feature Enhancer and the instance embeddings extracted from the keyframe.
Since the Feature Enhancer is the most computationally demanding component, bypassing the object decoder alone yields only limited latency reductions.
However, this strategy provides a strong upper bound on the temporal reusability of keyframe instance embeddings. The performance gap between the Reuse Objects and the full model represents the \textit{propagation drop}, incurred by reusing the instances detected on the keyframe for the inter-frames. The \textit{drop} from Reuse Objects to the other methods (LiteFlowNet2, RAFT, MPVSS, \ours) represents the cost of the specific architectural choices.

\tit{Optical Flow} Another common paradigm for mask propagation relies on off-the-shelf optical flow models. We evaluate two pre-trained networks, RAFT~\cite{teed2020raft} and LiteFlowNet2~\cite{hui20liteflownet2}, to compute the optical flow between consecutive frames and iteratively warp the initial masks generated by the slow network. For RAFT, we deliberately limit the inference to a single iteration; we observed that additional iterations yield negligible improvements in mask quality while severely penalizing the real-time efficiency which we aim to achieve. The RAFT model utilized is pre-trained on the Sintel~\cite{butler2012naturalistic,wulff2012lessons} and Flying Chairs~\cite{dosovitskiy2015flownet} datasets, whereas LiteFlowNet2 is pre-trained on the FlyingThings~\cite{mayer2016large} dataset.

\tit{MPVSS} We consider MPVSS~\cite{weng2023mask} as our primary competitor, as it is, to the best of our knowledge, the only existing method that proposes a dual-network propagation approach for object-centric segmentation models (specifically designed for fully supervised closed-set segmentation with Mask2Former~\cite{cheng2022masked}). To ensure a fair comparison, we trained MPVSS on top of all our adopted slow networks. We followed its original training recipe but restricted the training data to YouTubeVIS19, YouTubeVIS21, and OVIS, which are the video datasets present in our unified training corpus. Furthermore, we removed the classification loss required in their fully supervised setting. Similarly to SegFS, the MPVSS models were supervised strictly using mask and DICE losses. During inference, MPVSS conditions its optical flow prediction on the instance embeddings provided by the slow network to produce instance-wise optical flows. The keyframe masks are then iteratively warped using these object-specific flow fields.

\tit{Efficiency Analysis} To fully contextualize the performance of the models, the accuracy results in \cref{tab:model_comparison} must be analyzed in conjunction with the efficiency metrics detailed in \cref{tab:efficiency}. This table reports the FLOPs, the number of parameters, and inference latency across four edge devices and two GPUs. All efficiency metrics are computed assuming a $480 \times 480$ input and a vocabulary of 40 categories (mimicking the YouTubeVIS19 scenario), which primarily impacts the text-conditioning overhead of the slow networks.
Following MOBIUS, edge device latency was measured by compiling the models for NPU execution via the Qualcomm AI Hub. To capture diverse current and next-generation mobile hardware, we selected the Samsung Galaxy S25 Ultra (equipped with the Snapdragon 8 Elite for Galaxy), the Snapdragon 8 Gen 5 (powering upcoming devices such as the Samsung Galaxy S26, Xiaomi 17, OnePlus 15R, and Motorola Signature), the Snapdragon X2 Elite (targeting next-generation laptops), and the XR2 Gen 2 (utilized in the Meta Quest 3 headset). For GPU benchmarks, we report latency on NVIDIA A100 and T4 accelerators. For our proposed SegFS, the reported efficiency encompasses both the frozen backbone and the fast feature aggregator. Because the multi-scale features from the backbone are universally projected to a fixed channel dimension of 256, the computational contribution of the fast feature aggregator remains largely constant regardless of the chosen architecture. Consequently, any notable increases in latency, such as when employing the EfficientViT-L2, are inherently driven by the computational demands of the backbone itself and the initial projection step. From the amortized FPS column from \cref{tab:model_comparison}, we can observe that, on the MOBIUS-based slow models, \ours is the only approach able to cross the 30 FPS real-time threshold.

\tit{Performance Comparison} As demonstrated in \cref{tab:efficiency}, SegFS establishes itself as the most efficient fast path across all evaluated baselines. The MobileNetV4-CM variant of SegFS dominates almost all efficiency metrics, with the sole exception of latency on the NVIDIA A100, where the ResNet50 variant achieves the fastest execution. Beyond its efficiency, SegFS matches or surpasses the ``Reuse Objects'' baseline across several metrics and slow network configurations. This proves our core hypothesis: When the objective is strictly to propagate and reuse previously detected instances, the computationally heavy Feature Enhancer is redundant. Among the various slow network pairings, SegFS achieves its peak accuracy when built upon TROY-VIS. This strong synergy is a direct result of the TROY-VIS design philosophy, which allocates more representational power to the backbone while aggressively reducing the complexity of the Feature Enhancer. This architectural balance perfectly complements our fast path, explicitly relying on rich backbone features, though it inherently results in our heaviest and slowest SegFS configuration. Finally, we observe that MPVSS is highly portable from the fully supervised closed-set setting for which it was originally designed into our open-vocabulary scenario. However, despite its competitive accuracy, the on-device latency of MPVSS, along with the optical flow baselines RAFT and LiteFlowNet2, is severely bottlenecked by the heavy computational cost of mask warping, making these approaches fundamentally less viable for real-time mobile deployment than SegFS.

\begin{table}[t]
  \caption{Ablation study on the impact of our proposed model components, based on MOBIUS-Mini-M.}
  \vspace{-0.15cm}
  \label{tab:ablation}
  \centering
  \small 
  \setlength{\tabcolsep}{.35em}
  \resizebox{\textwidth}{!}{%
  \begin{tabular}{l cccc ccc ccc}
    \toprule
    & & & & \multicolumn{3}{c}{\textbf{YouTubeVIS19}} & \multicolumn{3}{c}{\textbf{OVIS}} \\
    \cmidrule(lr){5-7} \cmidrule(lr){8-10}
    Injection & Background Token & Attention Proj. & Smoothing & AP & AP$_{50}$ & AP$_{75}$ & AP & AP$_{50}$ & AP$_{75}$ \\
    \midrule
    \xmark & \xmark & \xmark & \xmark & 37.9 & 62.4 & 37.4 & 10.7 & 26.8 & 7.4\\
    \cmark & \xmark & \xmark & \xmark & 39.6 & 63.3 & 40.1 & 12.5 & 29.5 & 9.7\\
    \cmark & \cmark & \xmark & \xmark & 39.6 & 63.2 & 40.4 & 12.8 & 30.0 & 10.0\\
    \cmark & \cmark & \cmark & \xmark & 40.2 & 63.1 & 42.2 & 13.6 & 30.3 & 11.2\\
    \rowcolor{OurColor}
    \cmark & \cmark & \cmark & \cmark & 40.9 & 63.8 & 42.9 & 13.7 & 30.5 & 11.8\\
    \bottomrule
  \end{tabular}%
  }
\vspace{-0.3cm}
\end{table}

\begin{figure}[t]
    \centering
     \includegraphics[width=\linewidth]{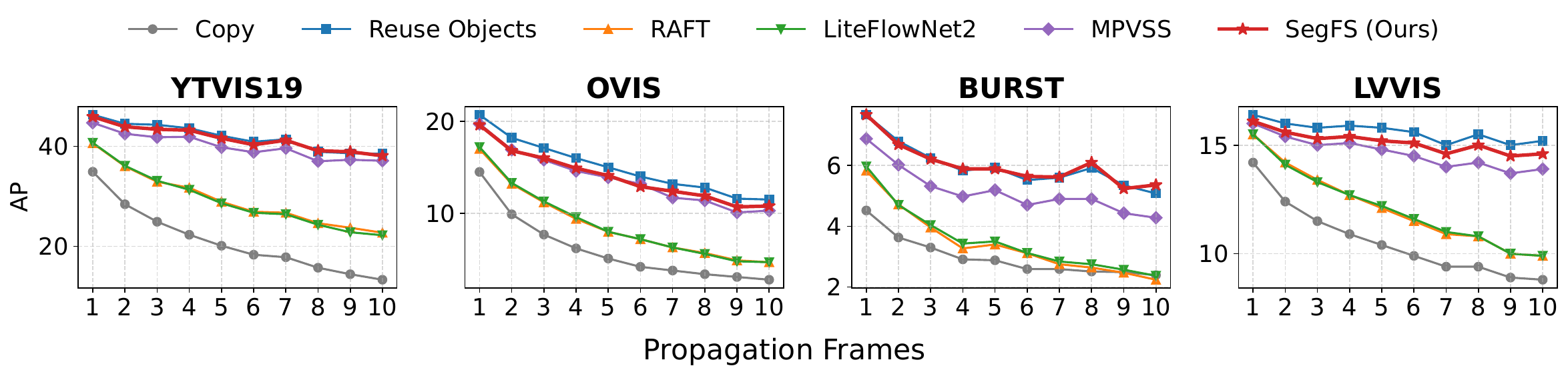}
     \vspace{-0.65cm}
    \caption{Propagation interval analysis. AP performance across datasets as the propagation interval ($T$) increases, evaluated with the MOBIUS-Mini-M slow network.
    }
    \label{fig:propagation_interval}
    \vspace{-0.6cm}
\end{figure}

\begin{figure}[t]
    \centering
     \includegraphics[width=\linewidth]{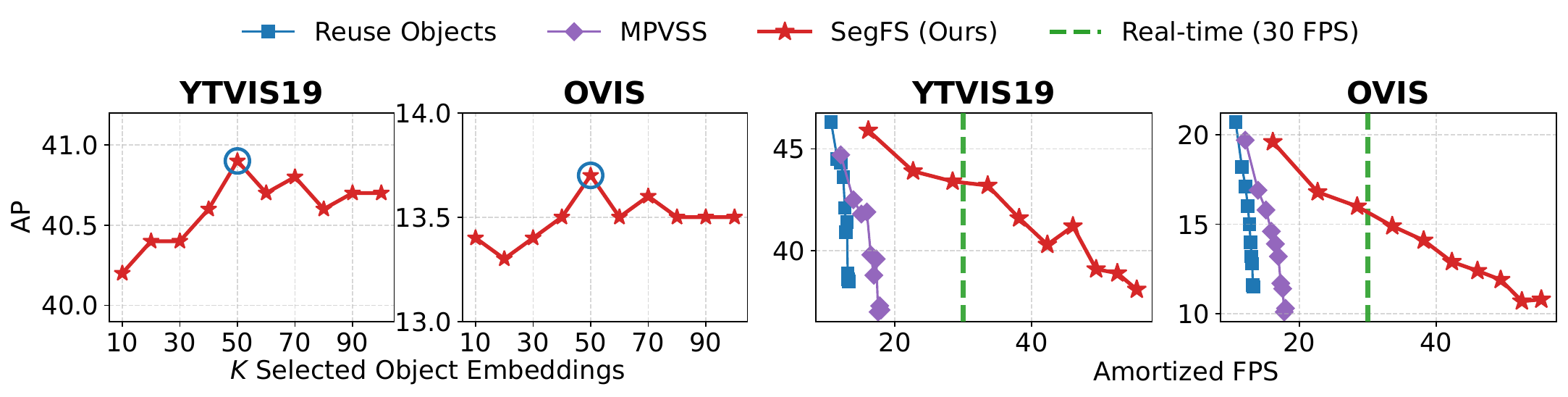}
     \vspace{-0.65cm}
    \caption{Ablation on the impact of $K$ on AP (left) and ablation on the impact of $T$ on AP and amortized FPS (right).
    }
    \label{fig:combined_k_T_ablation}
    \vspace{-0.3cm}
\end{figure}

\subsection{Ablation Studies and Analyses}

\tinytit{Propagation Interval} In \cref{fig:propagation_interval}, we analyze the impact of the mask propagation interval $T$ on the overall segmentation performance.
Using the MOBIUS model with the MobileNet V4-CM backbone as our slow network, we report the AP across all four evaluation datasets as $T$ increases from 1 to 10 frames.
As expected, performance universally declines as the propagation interval increases, as the keyframe outputs become less and less up-to-date%
, leading to spatial misalignments and tracking errors.
Furthermore, as $T$ grows, the quality of the keyframe results has less proportional impact on the overall metric.

The results show two distinct behavioral clusters
: \textit{(i)} approaches that propagate the semantic instance embeddings, namely ``Reuse Objects'', MPVSS, and our SegFS, demonstrate remarkable robustness, degrading much more gracefully as $T$ increases; \textit{(ii)} methods that rely on propagating or warping the pixel-level masks, such as ``Copy'', RAFT, and LiteFlowNet2, experience a severe and rapid drop in AP.
This observation validates our core intuition: Carrying forward the semantic object representations enables the fast network to consistently re-localize instances using current-frame visual features, effectively bypassing the compounding spatial errors inherent in direct mask warping.

Fig.~\ref{fig:combined_k_T_ablation} (left) reports AP vs.\ amortized FPS as $T$ varies in
$[1,10]$ on YTVIS19 and OVIS. SegFS dominates Reuse Objects and MPVSS across the entire range, and it is the only
configuration crossing the 30~FPS real-time threshold. For practical on-device deployment at 30~FPS acquisition (33~ms/frame), a more efficient strategy would be executing the Fast Feature Aggregator immediately after each frame is captured (with a latency of 8.3~ms) and overlaps the remaining $\sim$25~ms with the slow-path computation, yielding $T\!\approx\!5$--$6$.

\begin{figure}[t] 
    \centering
     \includegraphics[width=\linewidth]{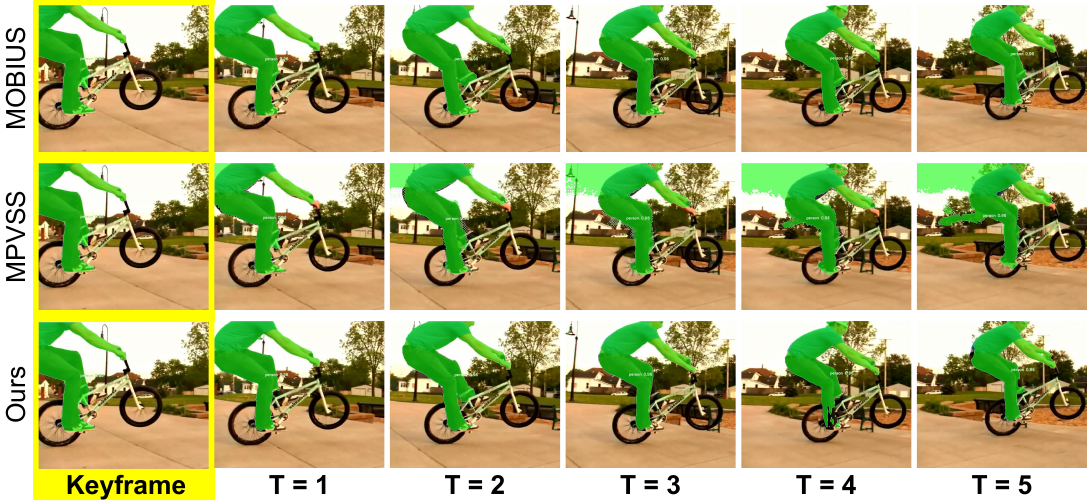}
    \vspace{-0.3cm}
    \caption{Resulting segmentations on a YouTubeVIS19 sequence, %
    showing the predictions of the slow network (MOBIUS-Mini-L); of MPVSS; and \ours (ours).}
    \label{fig:qualitatives_main}
    \vspace{-0.6cm}
\end{figure}

\tit{Model Components} To quantify the contribution of each architectural component within the fast path, we conduct a detailed ablation study in \cref{tab:ablation}.
For these experiments, we train the models for 100k iterations using the standard training recipe described in \cref{sec:setup}, employing the MOBIUS-Mini-M model (\ie, with a MobileNet V4-CM backbone) as our frozen slow network. The first row represents the simplest formulation of our fast network. In this configuration, we retain the progressive upsampling and the projection of the keyframe instance embeddings into the fast feature map. However, it does not include the object injection, the background token, the attention-based object projection, and the smoothing module. In the subsequent rows, we progressively add these missing components to evaluate their individual impact. The results demonstrate that the most significant performance jumps are driven by the introduction of the injection module and the attention-based prototype refinement. The injection module ensures that the fast network is explicitly conditioned by the rich semantic representations decoded by the slow network.
Furthermore, the attention-based refinement proves crucial, as it properly adapts the object queries before they are injected into the fast path, ultimately leading to our most accurate configuration.

\tit{Sensitivity Analysis on $K$} As in \cref{tab:ablation}, we trained SegFS with $K\!\in\!\{10, 20, \ldots, 100\}$ for 100k iterations on MOBIUS-Mini-M.
As shown in Fig.~\ref{fig:combined_k_T_ablation}, AP peaks at $K\!=\!50$ on both YTVIS19 and OVIS, then plateaus at a slightly lower value: beyond $K\!=\!50$, additional embeddings act as noise and no longer contribute to the Fast Feature Aggregator. $K$ controls two operations: \textit{(i)} the projection of the selected object embeddings to the $P_5$ feature space (slow path) and \textit{(ii)} the Object Guidance in the Fast Feature Aggregator (fast path). The cost of both is negligible across the entire range: From $K\!=\!10$ to $K\!=\!100$, the projection grows from 0.245 to 0.340~GFLOPs (0.324 to 0.444~ms) and the Fast Feature Aggregator from 6.323 to 6.333~GFLOPs (6.619 to 6.661~ms).

\tit{Qualitative Results} \cref{fig:qualitatives_main} shows qualitative results on YouTubeVIS19.
On the keyframe, we execute the slow model across all three configurations to extract the initial instance representations.
For the inter-frames, both \ours and MPVSS propagate the keyframe object embeddings.
MPVSS fails to accurately track the subject as they enter the scene from the top image boundary; optical flow inherently struggles to warp masks for newly introduced spatial content that was absent in previous frames.
As SegFS dynamically conditions the current-frame backbone features, it robustly re-localizes the instance regardless of motion, maintaining high-quality segmentation on par with running the heavy slow network on every frame.
\section{Conclusion}
\vspace{-2mm}
We introduced SegFS, a dual-path framework designed to overcome the computational bottlenecks of real-time OV-VIS on mobile devices. By projecting keyframe instance semantics directly into the intermediate backbone feature space, SegFS effectively bypasses the computationally expensive feature enhancer and pixel decoder. Experiments demonstrate that our approach achieves up to a 14× latency reduction over existing object-centric models, establishing a new standard for real-time, on-device OV-VIS while preserving competitive performance.

\FloatBarrier

\makeatletter\ifeccv@review\else
\subsection*{Acknowledgments}
Marcella Cornia acknowledges support from the EU Horizon project ``ELIAS - European Lighthouse of AI for Sustainability'' (GA No. 101120237).
\fi\makeatother
\clearpage

%
%
\bibliographystyle{splncs04}
\bibliography{main}

\begin{thebibliography}{10}
\providecommand{\url}[1]{\texttt{#1}}
\providecommand{\urlprefix}{URL }
\providecommand{\doi}[1]{https://doi.org/#1}

\bibitem{athar2023burst}
Athar, A., Luiten, J., Voigtlaender, P., Khurana, T., Dave, A., Leibe, B., Ramanan, D.: {BURST}: A benchmark for unifying object recognition, segmentation and tracking in video. In: WACV (2023)

\bibitem{butler2012naturalistic}
Butler, D., Wulff, J., Stanley, G., Black, M.J.: A naturalistic open source movie for optical flow evaluation. In: ECCV (2012)

\bibitem{cai2023efficientvit}
Cai, H., Li, J., Hu, M., Gan, C., Han, S.: {EfficientViT}: Lightweight multi-scale attention for high-resolution dense prediction. In: ICCV (2023)

\bibitem{carion2020end}
Carion, N., Massa, F., Synnaeve, G., Usunier, N., Kirillov, A., Zagoruyko, S.: End-to-end object detection with transformers. In: ECCV (2020)

\bibitem{cheng2022masked}
Cheng, B., Misra, I., Schwing, A., Kirillov, A., Girdhar, R.: Masked-attention mask transformer for universal image segmentation. In: CVPR (2022)

\bibitem{cheng2021per}
Cheng, B., Schwing, A., Kirillov, A.: Per-pixel classification is not all you need for semantic segmentation. In: NeurIPS (2021)

\bibitem{ding2022decoupling}
Ding, J., Xue, N., Xia, G.S., Dai, D.: Decoupling zero-shot semantic segmentation. In: CVPR (2022)

\bibitem{dosovitskiy2015flownet}
Dosovitskiy, A., Fischer, P., Ilg, E., Hausser, P., Hazirbas, C., Golkov, V., van~der Smagt, P., Cremers, D., Brox, T.: {FlowNet}: Learning optical flow with convolutional networks. In: ICCV (2015)

\bibitem{gupta2019lvis}
Gupta, A., Dollar, P., Girshick, R.: {LVIS}: A dataset for large vocabulary instance segmentation. In: CVPR (2019)

\bibitem{he2016deep}
He, K., Zhang, X., Ren, S., Sun, J.: Deep residual learning for image recognition. In: CVPR (2016)

\bibitem{Horn:1981:HS}
Horn, B., Schunck, B.: Determining optical flow. Artif. Intell.  \textbf{17}(1-3),  185--203 (1981)

\bibitem{Huang:2022:MinVIS}
Huang, D., Yu, Z., Anandkumar, A.: {MinVIS}: {A} minimal video instance segmentation framework without video-based training. In: NeurIPS (2022)

\bibitem{hui20liteflownet2}
Hui, T., Tang, X., Loy, C.C.: {A Lightweight Optical Flow CNN---Revisiting Data Fidelity and Regularization}. IEEE TPAMI  \textbf{43}(8),  2555--2569 (2020)

\bibitem{Ilg:2017:FNE}
Ilg, E., Mayer, N., Saikia, T., Keuper, M., Dosovitskiy, A., Brox, T.: {FlowNet} 2.0: Evolution of optical flow estimation with deep networks. In: CVPR (2017)

\bibitem{kirillov2023segment}
Kirillov, A., Mintun, E., Ravi, N., Mao, H., Rolland, C., Gustafson, L., Xiao, T., Whitehead, S., Berg, A., Lo, W.Y., et~al.: Segment anything. In: ICCV (2023)

\bibitem{krylov2021open}
Krylov, I., Nosov, S., Sovrasov, V.: {Open Images V5 Text Annotation and Yet Another Mask Text Spotter}. In: ACML (2021)

\bibitem{li2023mask}
Li, F., Zhang, H., Xu, H., Liu, S., Zhang, L., Ni, L., Shum, H.: Mask {DINO}: Towards a unified transformer-based framework for object detection and segmentation. In: CVPR (2023)

\bibitem{liang2023open}
Liang, F., Wu, B., Dai, X., Li, K., Zhao, Y., Zhang, H., Zhang, P., Vajda, P., Marculescu, D.: Open-vocabulary semantic segmentation with mask-adapted {CLIP}. In: CVPR (2023)

\bibitem{liang2025referdino}
Liang, T., Lin, K.Y., Tan, C., Zhang, J., Zheng, W., Hu, J.F.: {ReferDINO}: Referring video object segmentation with visual grounding foundations. In: ICCV (2025)

\bibitem{lin2014microsoft}
Lin, T., Maire, M., Belongie, S., Hays, J., Perona, P., Ramanan, D., Doll{\'a}r, P., Zitnick, L.: Microsoft {COCO}: Common objects in context. In: ECCV (2014)

\bibitem{liu2024grounding}
Liu, S., Zeng, Z., Ren, T., Li, F., Zhang, H., Yang, J., Jiang, Q., Li, C., Yang, J., Su, H., et~al.: Grounding {DINO}: Marrying {DINO} with grounded pre-training for open-set object detection. In: ECCV (2024)

\bibitem{Lucas:1981:AII}
Lucas, B., Kanade, T.: An iterative image registration technique with an application to stereo vision. In: IJCAI. pp. 674--679 (1981)

\bibitem{lv2024rt}
Lv, W., Zhao, Y., Chang, Q., Huang, K., Wang, G., Liu, Y.: {RT-DETRv2}: Improved baseline with bag-of-freebies for real-time detection transformer. arXiv:2407.17140 [cs.CV]  (2024)

\bibitem{mayer2016large}
Mayer, N., Ilg, E., Hausser, P., Fischer, P., Cremers, D., Dosovitskiy, A., Brox, T.: A large dataset to train convolutional networks for disparity, optical flow, and scene flow estimation. In: CVPR (2016)

\bibitem{nagaraja2016modeling}
Nagaraja, V., Morariu, V., Davis, L.: Modeling context between objects for referring expression understanding. In: ECCV (2016)

\bibitem{Nilsson:2018:SVS}
Nilsson, D., Sminchisescu, C.: Semantic video segmentation by gated recurrent flow propagation. In: CVPR (2018)

\bibitem{qi2022occluded}
Qi, J., Gao, Y., Hu, Y., Wang, X., Liu, X., Bai, X., Belongie, S., Yuille, A., Torr, P., Bai, S.: Occluded video instance segmentation: A benchmark. IJCV  \textbf{130}(8),  2022--2039 (2022)

\bibitem{qin2024mobilenetv4}
Qin, D., Leichner, C., Delakis, M., Fornoni, M., Luo, S., Yang, F., Wang, W., Banbury, C., Ye, C., Akin, B., et~al.: {MobileNetV4}: Universal models for the mobile ecosystem. In: ECCV (2024)

\bibitem{ravisam}
Ravi, N., Gabeur, V., Hu, Y.T., Hu, R., Ryali, C., Ma, T., Khedr, H., R{\"a}dle, R., Rolland, C., Gustafson, L., et~al.: {SAM} 2: Segment anything in images and videos. In: ICLR (2024)

\bibitem{roh2021sparse}
Roh, B., Shin, J., Shin, W., Kim, S.: Sparse {DETR}: Efficient end-to-end object detection with learnable sparsity. In: ICLR (2022)

\bibitem{segu2025mobius}
Segu, M., Gazulla, M.T., Xian, Y., Van~Gool, L., Tombari, F.: {MOBIUS}: Big-to-mobile universal instance segmentation via multi-modal bottleneck fusion and calibrated decoder pruning. In: ICCV (2025)

\bibitem{seo2020urvos}
Seo, S., Lee, J.Y., Han, B.: {URVOS}: Unified referring video object segmentation network with a large-scale benchmark. In: ECCV (2020)

\bibitem{Sevilla-Lara:2016:OFS}
Sevilla{-}Lara, L., Sun, D., Jampani, V., Black, M.J.: Optical flow with semantic segmentation and localized layers. In: CVPR (2016)

\bibitem{shao2019objects365}
Shao, S., Li, Z., Zhang, T., Peng, C., Yu, G., Zhang, X., Li, J., Sun, J.: {Objects365}: A large-scale, high-quality dataset for object detection. In: ICCV (2019)

\bibitem{teed2020raft}
Teed, Z., Deng, J.: {RAFT}: Recurrent all-pairs field transforms for optical flow. In: ECCV (2020)

\bibitem{wang2023towards}
Wang, H., Yan, C., Wang, S., Jiang, X., Tang, X., Hu, Y., Xie, W., Gavves, E.: Towards open-vocabulary video instance segmentation. In: ICCV (2023)

\bibitem{wang2021unidentified}
Wang, W., Feiszli, M., Wang, H., Tran, D.: Unidentified video objects: A benchmark for dense, open-world segmentation. In: ICCV (2021)

\bibitem{weng2023mask}
Weng, Y., Han, M., He, H., Li, M., Yao, L., Chang, X., Zhuang, B.: Mask propagation for efficient video semantic segmentation. In: NeurIPS (2023)

\bibitem{wu2024general}
Wu, J., Jiang, Y., Liu, Q., Yuan, Z., Bai, X., Bai, S.: General object foundation model for images and videos at scale. In: CVPR (2024)

\bibitem{wulff2012lessons}
Wulff, J., Butler, D., Stanley, G., Black, M.J.: Lessons and insights from creating a synthetic optical flow benchmark. In: ECCV (2012)

\bibitem{Xiao:2018:MND}
Xiao, H., Feng, J., Lin, G., Liu, Y., Zhang, M.: {MoNet}: Deep motion exploitation for video object segmentation. In: CVPR (2018)

\bibitem{xu2022simple}
Xu, M., Zhang, Z., Wei, F., Lin, Y., Cao, Y., Hu, H., Bai, X.: A simple baseline for open-vocabulary semantic segmentation with pre-trained vision-language model. In: ECCV (2022)

\bibitem{UNINEXT}
Yan, B., Jiang, Y., Wu, J., Wang, D., Yuan, Z., Luo, P., Lu, H.: Universal instance perception as object discovery and retrieval. In: CVPR (2023)

\bibitem{yan2025towards}
Yan, B., Sundermeyer, M., Tan, D.J., Lu, H., Tombari, F.: Towards real-time open-vocabulary video instance segmentation. In: WACV (2025)

\bibitem{yang2019video}
Yang, L., Fan, Y., Xu, N.: Video instance segmentation. In: ICCV (2019)

\bibitem{yao2021efficientdetr}
Yao, Z., Ai, J., Li, B., Zhang, C.: Efficient {DETR}: Improving end-to-end object detector with dense prior. arXiv:2104.01318 [cs.CV]  (2021)

\bibitem{yu2020bdd100k}
Yu, F., Chen, H., Wang, X., Xian, W., Chen, Y., Liu, F., Madhavan, V., Darrell, T.: {BDD100K}: A diverse driving dataset for heterogeneous multitask learning. In: CVPR (2020)

\bibitem{zang2022open}
Zang, Y., Li, W., Zhou, K., Huang, C., Loy, C.C.: Open-vocabulary {DETR} with conditional matching. In: ECCV (2022)

\bibitem{zhang2023mobilesamv2}
Zhang, C., Han, D., Zheng, S., Choi, J., Kim, T., Hong, C.S.: {MobileSAMv2}: Faster segment anything to everything. arXiv:2312.09579 [cs.CV]  (2023)

\bibitem{zhang2023simple}
Zhang, H., Li, F., Zou, X., Liu, S., Li, C., Yang, J., Zhang, L.: A simple framework for open-vocabulary segmentation and detection. In: ICCV (2023)

\bibitem{Zhang:2024:MobileInst}
Zhang, R., Cheng, T., Yang, S., Jiang, H., Zhang, S., Lyu, J., Li, X., Ying, X., Gao, D., Liu, W., Wang, X.: {MobileInst}: Video instance segmentation on the mobile. In: AAAI (2024)

\bibitem{zhao2023fast}
Zhao, X., Ding, W., An, Y., Du, Y., Yu, T., Li, M., Tang, M., Wang, J.: Fast segment anything. arXiv:2306.12156 [cs.CV]  (2023)

\bibitem{zhao2024detrs}
Zhao, Y., Lv, W., Xu, S., Wei, J., Wang, G., Dang, Q., Liu, Y., Chen, J.: {DETR}s beat {YOLO}s on real-time object detection. In: CVPR (2024)

\bibitem{zheng2023less}
Zheng, D., Dong, W., Hu, H., Chen, X., Wang, Y.: Less is more: Focus attention for efficient {DETR}. In: ICCV (2023)

\bibitem{zhu2021deformable}
Zhu, X., Su, W., Lu, L., Li, B., Wang, X., Dai, J.: Deformable {DETR}: Deformable transformers for end-to-end object detection. In: ICLR (2021)

\bibitem{zou2023generalized}
Zou, X., Dou, Z.Y., Yang, J., Gan, Z., Li, L., Li, C., Dai, X., Behl, H., Wang, J., Yuan, L., et~al.: Generalized decoding for pixel, image, and language. In: CVPR (2023)

\bibitem{zou2023segment}
Zou, X., Yang, J., Zhang, H., Li, F., Li, L., Wang, J., Wang, L., Gao, J., Lee, Y.J.: Segment everything everywhere all at once. In: NeurIPS (2023)

\end{thebibliography}
\clearpage
\appendix
\title{Segmenting, Fast and Slow:\texorpdfstring{\\}{ }%
Real-Time Open-Vocabulary Video Instance Segmentation with Dual-Path Processing}
\titlerunning{Supplementary Material}

\makesupptitle 

In this supplementary material, we extend the main paper as follows: (i) in \cref{sec:additional_training_details}, we report additional details for the reproducibility of our method, including further training details and information regarding the employed datasets; (ii) in \cref{sec:additional_component_analysis}, we extend the efficiency analysis by studying the computational behavior of the proposed fast path across different backbone architectures and vocabulary sizes; (iii) in \cref{sec:additional_qualitative}, we present additional qualitative examples, showing comparisons with competing propagation strategies and illustrating the ability of our model to segment multiple instances and small objects across diverse scenarios.

\section{Additional Training Details}
\label{sec:additional_training_details}

\begin{table}[!b]
\caption{Training datasets. For each dataset, we report the number of images, the number of annotated objects, the type of supervision provided (category labels or referring expressions), and the sampling ratio used during training.}
\label{tab:datasets_ratios}
\centering
\begin{tabular}{lcccc}
\toprule
\textbf{Dataset} & \textbf{\# Images} & \textbf{\# Objects} & \textbf{Annotations} & \textbf{Samp. Ratio} \\
\midrule
\rowcolor{gray!15} \multicolumn{5}{l}{\textbf{Detection Data}} \\
LVIS~\cite{gupta2019lvis} & 100170 & 1270141 & Category & 1.5 \\
COCO~\cite{lin2014microsoft} & 118287 & 860001 & Category & 1.5 \\
BDD~\cite{yu2020bdd100k} & 69863 & 1274792 & Category & 0.15 \\
\rowcolor{gray!15} \multicolumn{5}{l}{\textbf{Grounding Data}} \\
RefCOCO~\cite{nagaraja2016modeling} & 16994 & 42404 & Expression & 2.5 \\
RefCOCOg~\cite{nagaraja2016modeling} & 21899 & 42226 & Expression & 2.5 \\
RefCOCO+~\cite{nagaraja2016modeling} & 77396 & 3596689 & Expression & 2.5 \\
\rowcolor{gray!15} \multicolumn{5}{l}{\textbf{Open-World Data}} \\
UVO~\cite{wang2021unidentified} & 16923 & 157624 & - & 0.2 \\
SA1B~\cite{kirillov2023segment} & 2147712 & 99427126 & - & 2.5 \\
\rowcolor{gray!15} \multicolumn{5}{l}{\textbf{Video Data}} \\
YTVIS19~\cite{yang2019video} & 61845 & 97110 & Category & 0.3 \\
YTVIS21~\cite{yang2019video} & 90160 & 175384 & Category & 0.3 \\
OVIS~\cite{qi2022occluded} & 42149 & 206092 & Category & 0.3 \\
RVOS~\cite{seo2020urvos} & 93857 & 159961 & Expression & 0.3 \\
\bottomrule
\end{tabular}
\end{table}

In \cref{sec:setup}, we introduced our main training details, including the list of employed datasets, following the unified single stage proposed in MOBIUS~\cite{segu2025mobius}. In \cref{tab:datasets_ratios}, we report additional details on the training datasets, including the number of images, annotated objects, type of annotation, and sampling ratio. Following GLEE~\cite{wu2024general} and MOBIUS, we apply mask IoU–based NMS, using mask area as the scoring metric, to filter out part-level annotations and retain object-level instances in SA-1B~\cite{kirillov2023segment}. During the 500,000 training iterations, we use the AdamW optimizer with a learning rate of 1e-4 and a weight decay of 0.05 that increases to 0.1 after 400,000 iterations. We employ a multi-scale augmentation and resize the input images such that the shortest side is at least 320 and at most 640.

\section{Additional Component Analysis}
\label{sec:additional_component_analysis}

\begin{figure}[!t]
    \centering
     \includegraphics[width=\linewidth]{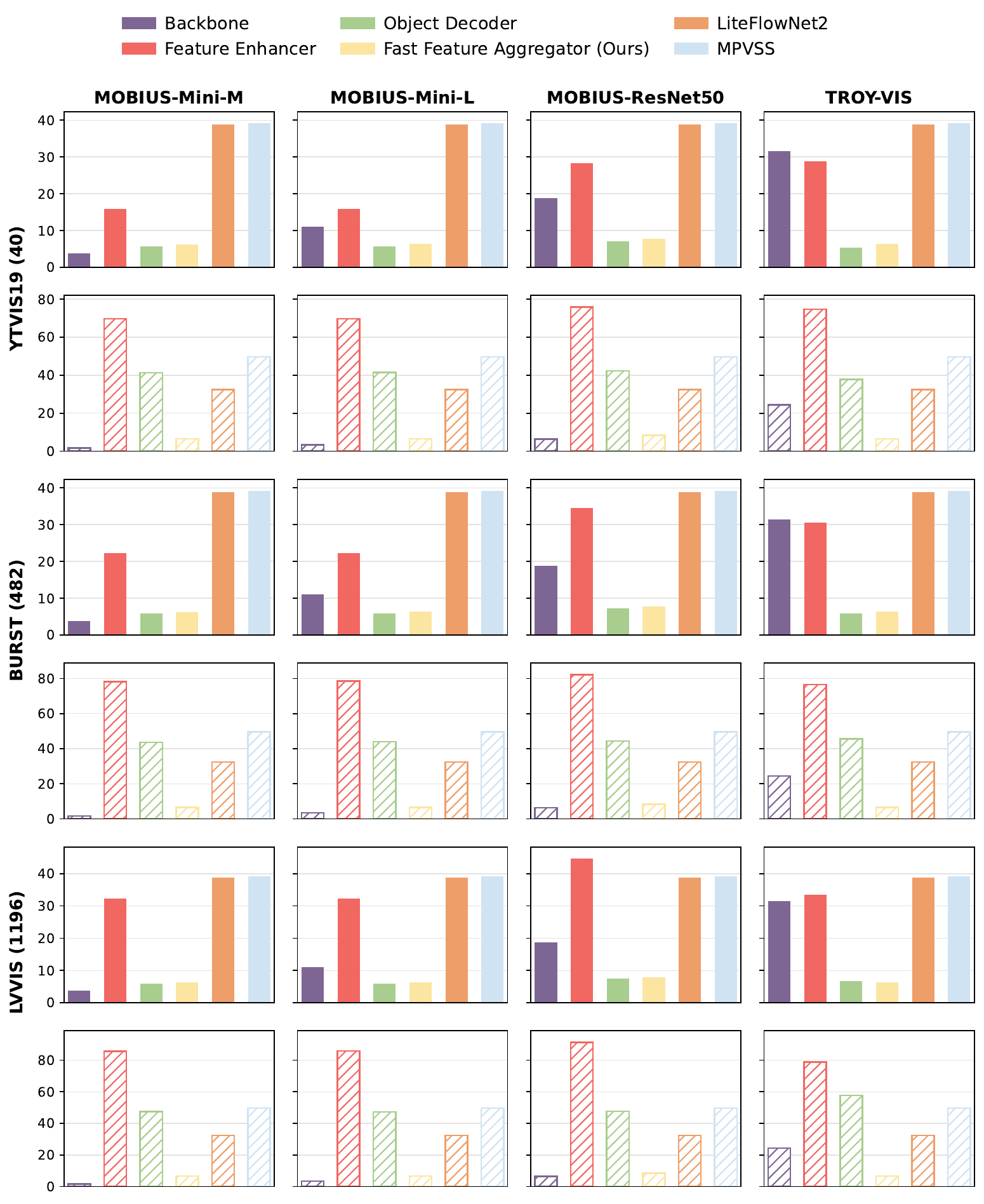}
    \vspace{-0.5cm}
    \caption{Efficiency analysis of the OV-VIS architectures. We report the FLOPs (full bars) and on-device latency on a Samsung Galaxy S25 Ultra (striped bars) for the core components of four baseline models, MOBIUS-Mini-M,  MOBIUS-Mini-L,  MOBIUS-ResNet50~\cite{segu2025mobius}, and TROY-VIS~\cite{wang2023towards}, when assuming an input image with resolution 480$\times$480 and varying the number of categories (40 as in YouTubeVIS19, 482 as in BURST, and 1196 as in LVVIS), compared with Fast Feature Aggregation, from our proposed SegFS, LiteFlowNet2, and MPVSS.}
    \label{fig:flops_and_latency_histogram_supp}
    \vspace{-0.3cm}
\end{figure}

\cref{fig:flops_and_latency_histogram_supp} extends the efficiency analysis presented in \cref{fig:flops_and_latency_histogram} and \cref{tab:efficiency} of the main paper. Recall that a full OV-VIS pipeline consists of the backbone, the feature enhancer, and the object decoder, while our approach replaces the feature enhancer and object decoder with the proposed Fast Feature Aggregator (SegFS), requiring only the backbone of the corresponding OV-VIS model and our lightweight module. As shown in \cref{fig:flops_and_latency_histogram_supp}, the computational cost introduced by our Fast Feature Aggregator, as well as by LiteFlowNet2~\cite{hui20liteflownet2} and MPVSS~\cite{weng2023mask}, remains constant across different backbone architectures and numbers of categories. In contrast, the computational cost of the Feature Enhancer increases with the number of categories, making the full OV-VIS models progressively more expensive as the vocabulary size grows.
From this analysis, we observe remarkable latency improvements on a Samsung Galaxy S25 Ultra. When considering 40 categories (as in YouTubeVIS19~\cite{yang2019video}), our approach is approximately $14\times$, $12\times$, $7\times$, and $4\times$ faster than the full MOBIUS-Mini-M, MOBIUS-Mini-L, MOBIUS-ResNet50~\cite{segu2025mobius}, and TROY-VIS~\cite{wang2023towards} models, respectively. The speedup further increases when scaling to 1196 categories (LVVIS~\cite{wang2023towards}), reaching $16\times$, $14\times$, $9\times$, and $5\times$ for the same models.

We further clarify two aspects of the efficiency analysis. First, we do not include the computational cost of the text encoder when reporting FLOPs and latency. Following the assumptions adopted in MOBIUS and the caching strategy proposed in TROY-VIS, we assume that the set of textual categories remains fixed throughout the processing of a video. Under this assumption, the corresponding text embeddings can be pre-computed once before inference and reused for all frames of the sequence. Consequently, the text encoder does not contribute to the per-frame inference cost and is omitted from the efficiency analysis.

Second, the projection of the object embeddings to the fast feature space is executed on keyframes within the slow path. This projection introduces a small additional computational overhead that is not included in the efficiency comparison reported in the main paper. However, this cost is negligible in practice. When instantiated on top of the MOBIUS-Mini-M configuration, the projection module adds only 0.29 GFLOPs, 5.1M parameters, and 0.356 ms of latency on a Samsung Galaxy S25 Ultra. Given that this operation is executed only on sparse keyframes and represents a very small fraction of the total inference cost, we consider its impact negligible for the purposes of the efficiency analysis.

\clearpage
\section{Additional Qualitative Examples}
\label{sec:additional_qualitative}

\tit{Comparison with other methods}
We refer to \cref{fig:additional_qualitatives_pt1} and \cref{fig:additional_qualitatives_pt2} of the supplementary, each illustrating two representative video sequences. The sequences in \cref{fig:additional_qualitatives_pt1} depict (i) a lion cub being petted by a human hand and (ii) a hand releasing a music box and leaving the scene. \cref{fig:additional_qualitatives_pt2} shows (i) a swan diving into water and (ii) a Rubik’s cube placed next to a speed-stacking timer.
Across all sequences, the qualitative comparison highlights the limitations of simple mask propagation strategies. The \textit{Copy} baseline exhibits a strong ghosting effect, as the masks predicted on the keyframe are directly reused for the subsequent frames without accounting for motion or appearance changes. This setting effectively simulates what happens when only the slow model is executed in a real-time scenario and cannot provide updated segmentations at the same frame rate as the acquisition system, causing the masks to remain frozen over the intermediate frames.
Optical-flow-based propagation, represented by MPVSS~\cite{weng2023mask}, alleviates this issue but remains sensitive to large appearance variations and to objects entering or leaving the scene. For example, in the swan sequence (\cf \cref{fig:additional_qualitatives_pt2}), MPVSS struggles when the instance changes its pose while diving into the water. Similarly, in the other sequences involving human hands, the method produces noisy masks when new objects appear or disappear from the frame.
Our method exhibits a behaviour similar to the \textit{Reuse Objects} baseline, namely the ability to remain robust under significant scene variations and to re-detect the objects from the current frame. However, unlike \textit{Reuse Objects}, which still requires running the expensive feature enhancer of the slow model, our approach achieves this behaviour with substantially lower latency by relying on the lightweight Fast Feature Aggregator.

\begin{figure}[!t]
    \centering
     \includegraphics[width=\linewidth]{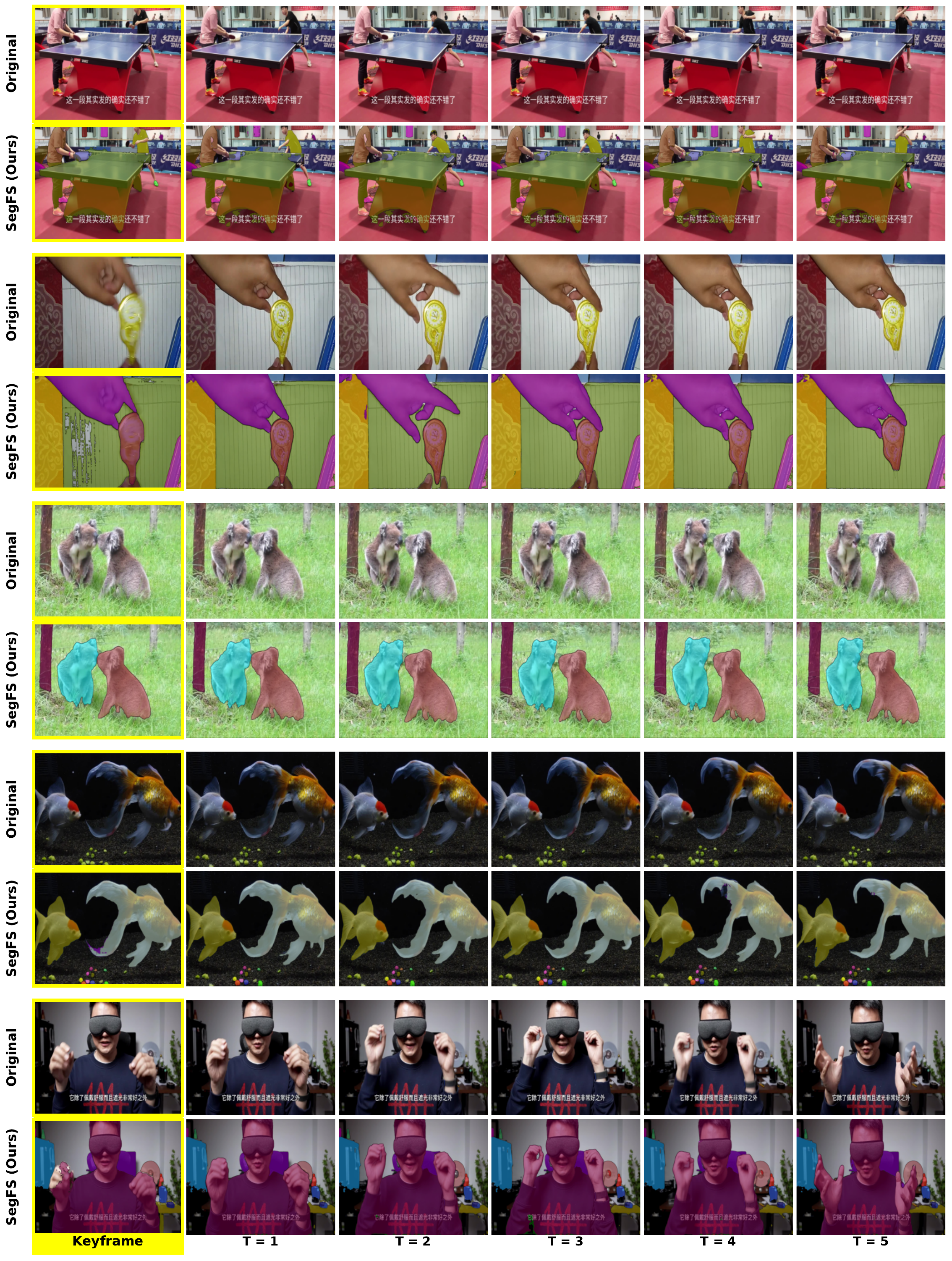}
    \vspace{-0.5cm}
    \caption{Additional qualitative results of \ours on five diverse video sequences from LVVIS, showing the original frames and our predictions, when using the MOBIUS-Mini-L as the slow model.}
    \label{fig:additional_qualitatives_pt3}
    \vspace{-0.3cm}
\end{figure}

\tit{Multiple and Small Instances}
\cref{fig:additional_qualitatives_pt3} presents five additional video sequences, showing the original frames along the predictions of our method. These examples highlight several important properties of the proposed approach. First, our model is able to consistently re-identify multiple instances belonging to the same category. This observation supports our intuition that the multi-scale backbone feature maps already contain sufficient spatial and semantic information to support instance segmentation when properly conditioned. For example, the model correctly distinguishes the two koalas in the third sequence and the two fish in the fourth sequence.
Second, our method remains effective even in scenes containing numerous small instances. This behaviour can be observed in the first sequence, where multiple shoes are segmented individually, and in the fourth sequence, where several small rocks are correctly identified as distinct instances.
Finally, the fast path is able to maintain stable segmentations over time and, in some cases, even refine imperfect masks predicted by the slow model on the keyframe. For instance, the shirt of the background player in the first sequence is consistently segmented across frames, while the background region in the second sequence and the hands of the person in the fifth sequence are improved with respect to the slow model on the keyframe.

\begin{figure}[t]
    \centering

    \begin{subfigure}{\textwidth}
        \centering
        \includegraphics[width=\textwidth]{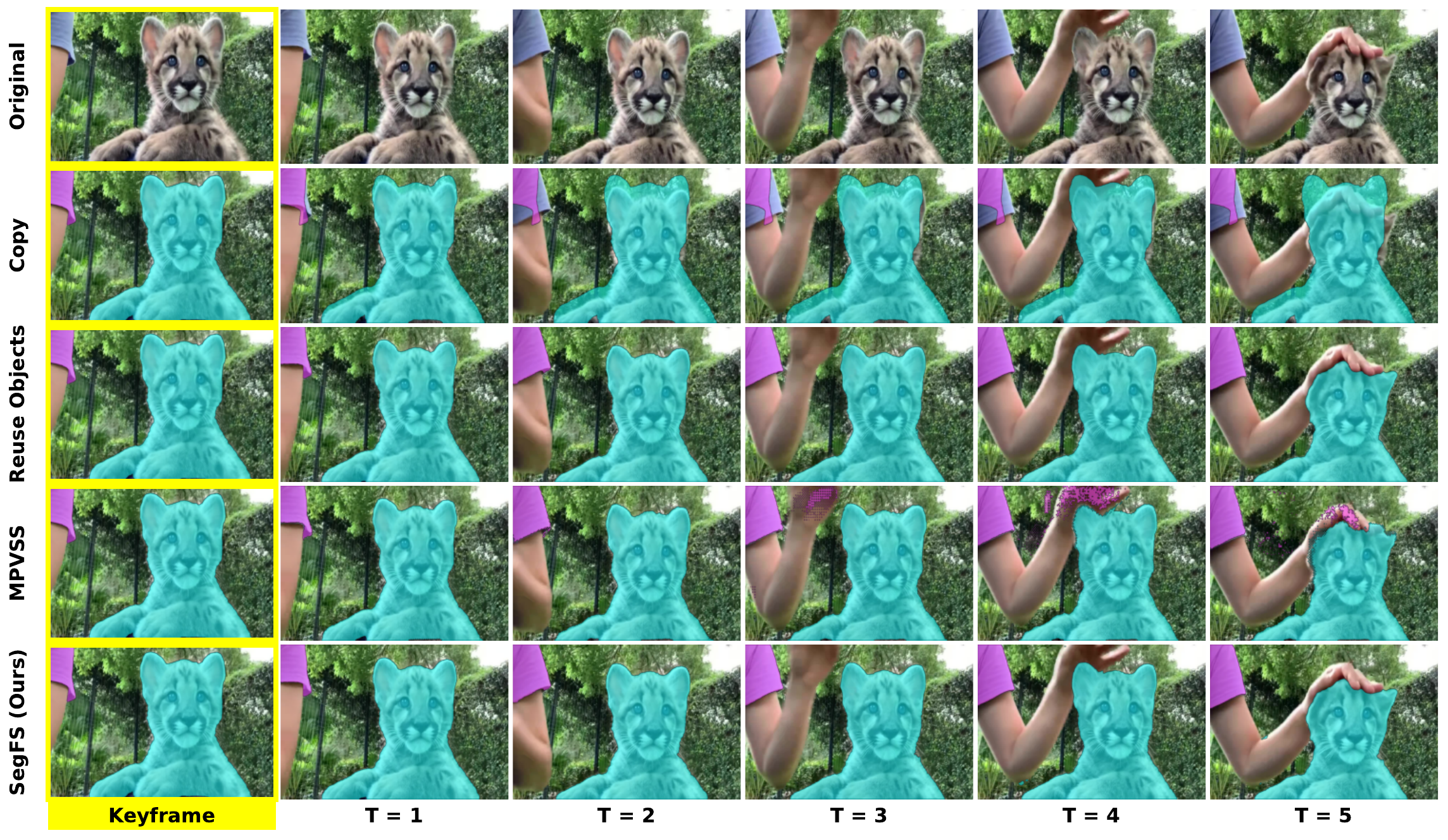}
        \label{fig:top}
    \end{subfigure}

    \vspace{0.5em}

    \begin{subfigure}{\textwidth}
        \centering
        \includegraphics[width=\textwidth]{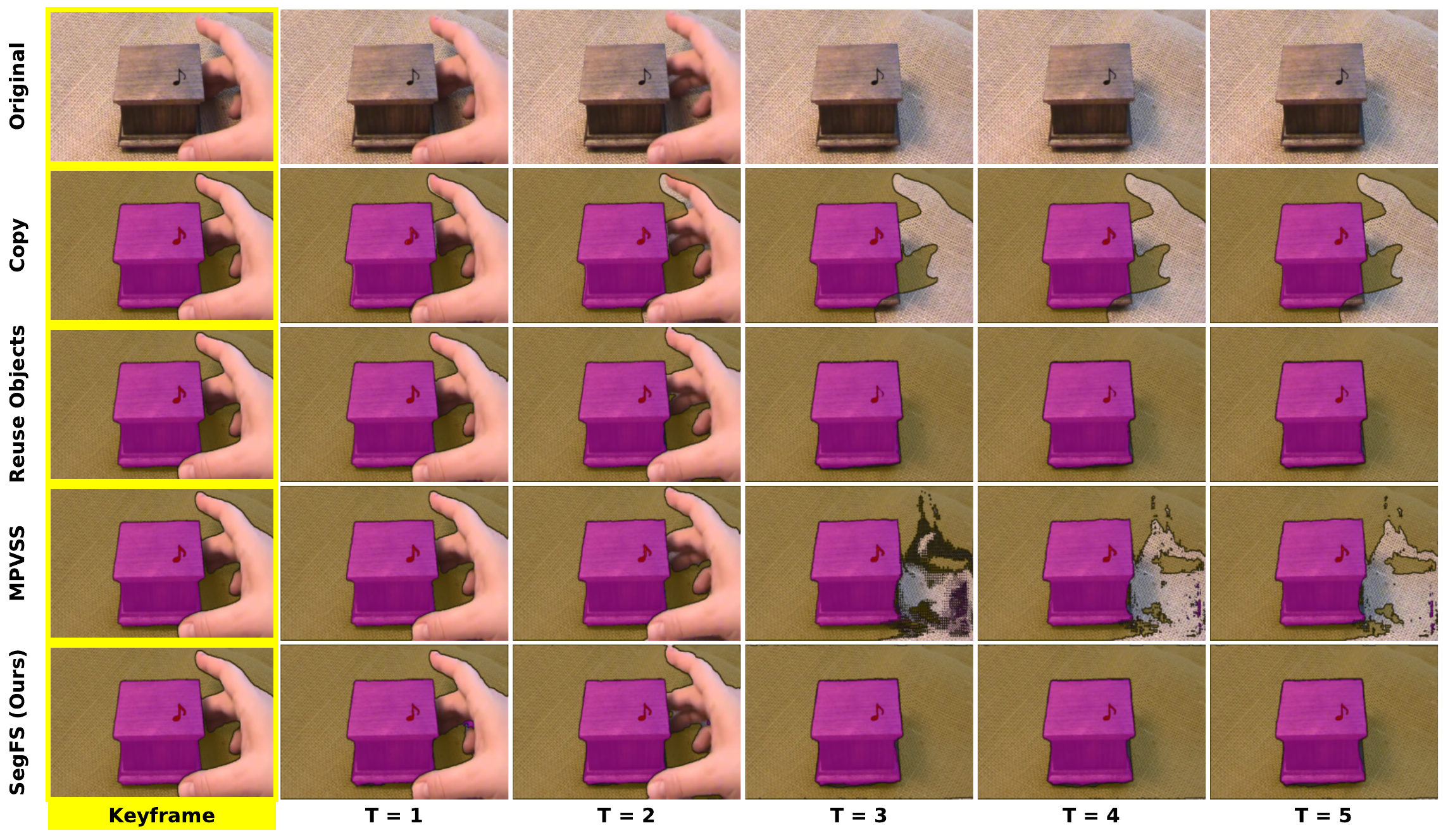}
        \label{fig:bottom2}
    \end{subfigure}
    \vspace{-0.8cm}
    \caption{Resulting segmentations on two LVVIS video sequences, showing the original frames, and predictions of the \textit{Copy} and \textit{Reuse Objects} approaches, of MPVSS, and \ours (ours), when employing the MOBIUS-Mini-L as the slow model.}
    \label{fig:additional_qualitatives_pt1}
\end{figure}

\begin{figure}[t]
    \centering

    \begin{subfigure}{\textwidth}
        \centering
        \includegraphics[width=\textwidth]{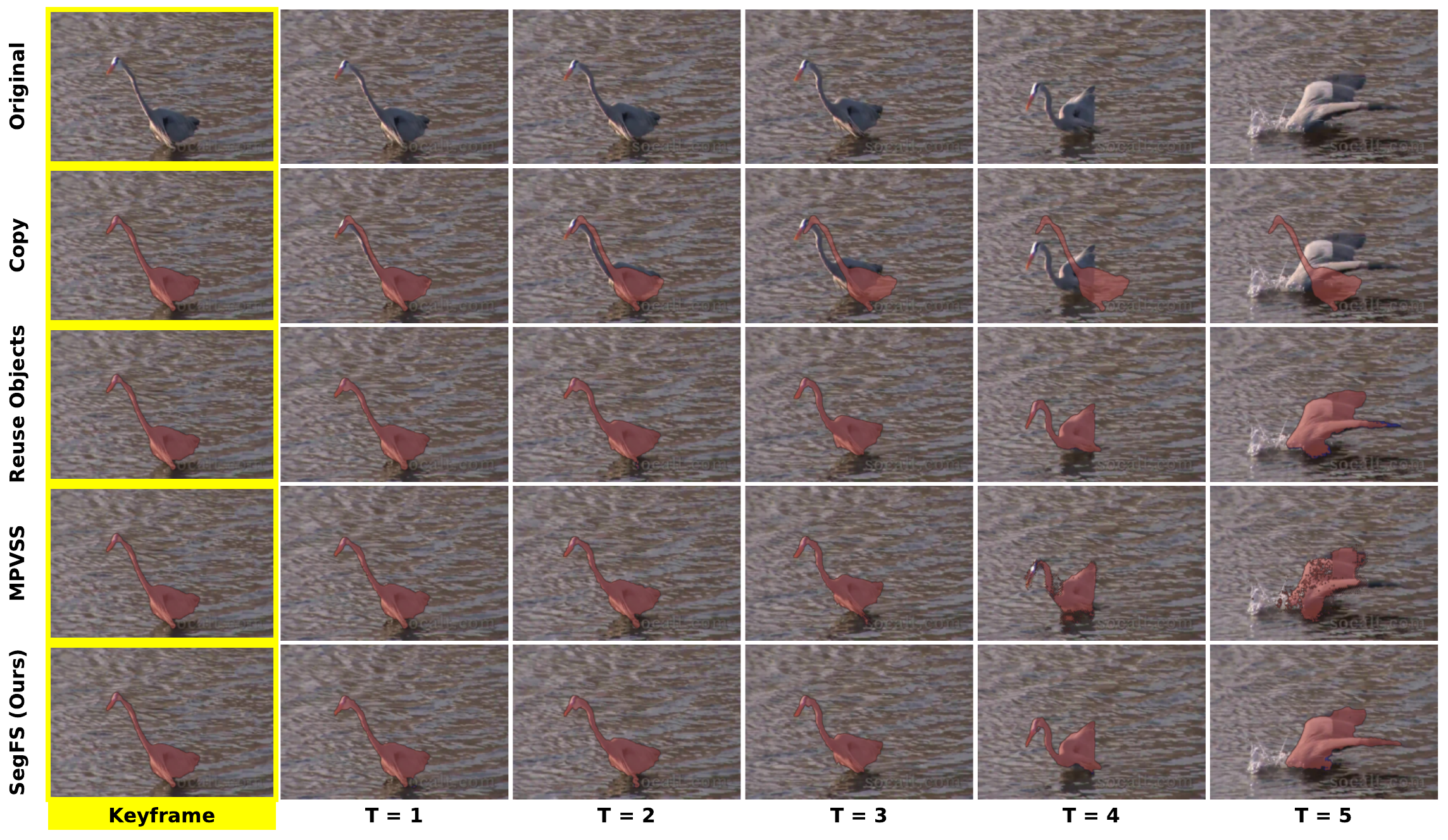}
        \label{fig:top2}
    \end{subfigure}

    \vspace{0.5em}

    \begin{subfigure}{\textwidth}
        \centering
        \includegraphics[width=\textwidth]{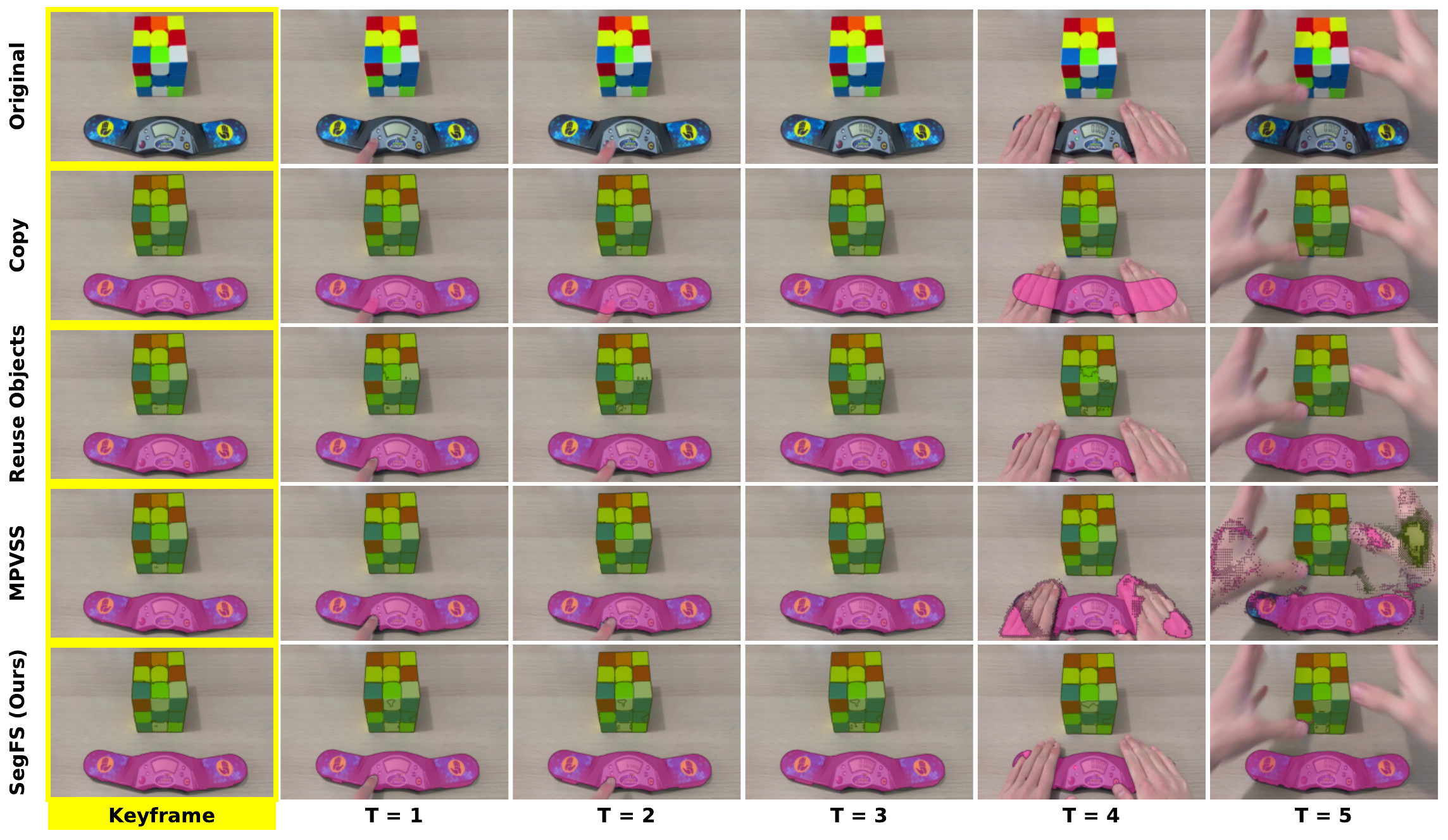}
        \label{fig:bottom}
    \end{subfigure}
    \vspace{-0.8cm}
    \caption{Additional resulting segmentations on two LVVIS video sequences, showing the original frames, and predictions of the \textit{Copy} and \textit{Reuse Objects} approaches, of MPVSS, and \ours (ours), when employing the MOBIUS-Mini-L as the slow model.}
    \label{fig:additional_qualitatives_pt2}
\end{figure}

\end{document}